%% file: main.tex
\documentclass[sigconf]{acmart}

\usepackage{amsmath,amsfonts}
\usepackage{algorithmic}
\usepackage{graphicx}
\usepackage{textcomp}
\usepackage{xspace}
\usepackage{xcolor}
\usepackage{soul}
\usepackage{pifont}
\usepackage{url}
\usepackage{hyperref}
\usepackage{booktabs}
\usepackage{multirow}
\usepackage{enumitem}
\usepackage{subcaption}
\usepackage{comment}
\usepackage{mathtools}

\newcommand{\vect}[1]{\boldsymbol{#1}}
\newcommand{\cardinality}[1]{\left| #1\right|}

\soulregister\cite7
\soulregister\ref7
\soulregister\pageref7

\AtBeginDocument{%
  }

\copyrightyear{2026}
\acmYear{2026}
\setcopyright{cc}
\setcctype{by}
\acmConference[ICSE '26]{2026 IEEE/ACM 48th International Conference on Software Engineering}{April 12--18, 2026}{Rio de Janeiro, Brazil}
\acmBooktitle{2026 IEEE/ACM 48th International Conference on Software Engineering (ICSE '26), April 12--18, 2026, Rio de Janeiro, Brazil}
\acmPrice{}
\acmDOI{10.1145/3744916.3773190}
\acmISBN{979-8-4007-2025-3/2026/04}

\newcommand{\approach}{{\sc{\texttt{MODA}}}\xspace}

\newcommand{\circled}[1]{\raisebox{-0.2ex}{\scalebox{1.1}{\ding{\numexpr201+#1\relax}}}}

\begin{document}

\title{DNN Modularization via Activation-Driven Training}

\author{Tuan Ngo}
\orcid{0000-0001-7136-7529}
\affiliation{%
  \institution{University of Southern California}
  \city{Los Angeles}
  \state{CA}
  \country{USA}
}
\email{tkngo@usc.edu}

\author{Abid Hassan}
\orcid{0009-0006-1022-4907}
\affiliation{%
  \institution{University of Southern California}
  \city{Los Angeles}
  \state{CA}
  \country{USA}
}
\email{mdskabid@usc.edu}

\author{Saad Shafiq}
\orcid{0000-0002-5901-1420}
\affiliation{%
  \institution{University of Southern California}
  \city{Los Angeles}
  \state{CA}
  \country{USA}
}
\email{sshafiq@usc.edu}

\author{Nenad Medvidovi\'c}
\orcid{0000-0002-1906-4878}
\affiliation{%
  \institution{University of Southern California}
  \city{Los Angeles}
  \state{CA}
  \country{USA}
}
\email{neno@usc.edu}

\renewcommand{\shortauthors}{Ngo et al.}

\input{sections/0.abstract}



\maketitle

\input{sections/1.introduction}

\input{sections/2.background}
\input{sections/3.approach}

\input{sections/4.evaluation}

\input{sections/5.results}

\input{sections/7.threats}
\input{sections/8.related_work}
\input{sections/6.discussion}

\clearpage
\bibliographystyle{ACM-Reference-Format}
\bibliography{references}

\end{document}

%% file: sections/0.abstract.tex
\begin{abstract}

\looseness-1
Deep Neural Networks (DNNs) tend to accrue technical debt and suffer from significant retraining costs when adapting to evolving requirements.
Modularizing DNNs offers the promise of improving their reusability. Previous work has proposed techniques to decompose DNN models into modules both during and after training. However, these strategies yield several shortcomings, including significant weight overlaps and accuracy losses across modules, restricted focus on convolutional layers only, and added complexity and training time by introducing auxiliary masks to control modularity. %
In this work, we propose \approach, an activation-driven modular training approach. \approach promotes inherent modularity within a DNN model by directly regulating the activation outputs of its layers based on three modular objectives: intra-class affinity, inter-class dispersion, and compactness. \approach is evaluated using three well-known DNN models %
and five datasets %
with varying sizes. This evaluation indicates that, compared to the existing state-of-the-art, using \approach yields several advantages: \textit{(1)}~\approach accomplishes modularization with 22\% less training time; \textit{(2)}~the resultant modules generated by \approach comprise up to 24x fewer weights and 37x less weight overlap while \textit{(3)}~preserving the original model's accuracy without additional fine-tuning; in module replacement scenarios, \textit{(4)}~\approach improves the accuracy of a target class by 12\% on average while ensuring minimal impact on the accuracy of other classes.

\end{abstract}

%% file: sections/1.introduction.tex
\section{Introduction}

\looseness-1
Deep neural networks (DNNs) have demonstrated exceptional capabilities in a range of domains (e.g.,~\cite{alexnetmodel,nassif2019speech,collobert2011natural}). Typically, DNN models are delivered as monolithic packages with numerous learned parameters tailored for particular tasks~\cite{alexnetmodel, vggmodel, resnetmodel, mobilenetmodel}.
Similar to monolithic software code, DNNs are prone to technical debt; specifically, the high inter-connectivity in their parameter space causes \emph{entanglement debt}~\cite{sculley2015hidden}. This debt can %
limit a DNN's adaptability to changing requirements. Approaches such as transfer learning~\cite{zhuang2020comprehensive} and continual learning~\cite{wang2024comprehensive} allow the reuse of learned model parameters and construction of new models for emerging requirements. However, %
these approaches still mandate the reuse of entire originating DNN models, even for tasks that require only a fraction of their functionalities. This incurs unnecessary memory and computation overhead~\cite{qi2023reusing}. 
Furthermore, these approaches suffer from catastrophic forgetting~\cite{mccloskey1989catastrophic, wang2024comprehensive}, where adapting a model to accommodate additional tasks may degrade its previously learned tasks.
This can cause a detrimental cascading effect, where the model must be repeatedly trained to adapt to changing requirements.

\looseness-1
To help address this, DNN modularization~\cite{pan2020decomposing, pan2022decomposing, imtiaz2023decomposing, cnnsplitter, qi2023reusing, gradsplitter, incite, mwt, ren2023deeparc} has recently emerged as a promising direction for enhancing model reuse.
Inspired by modularizing traditional software~\cite{parnas1972criteria, parnas1976design},
DNN modularization aims to decompose an $n$-class classification model into $n$ distinct groups of weights (i.e., modules), with each module dedicated to recognizing a single output class.
This offers the promise of %
selective integration of \textit{reusable} modules to assemble new DNN models with \textit{minimal to no retraining} required~\cite{gradsplitter, mwt}. 
Moreover, decomposing a DNN model into modules would simplify the \textit{removal} of unwanted classes and enable  \textit{replacement} of under-performing modules for specific classes with more accurate ones.

Research in DNN modularization has taken two primary directions: post-training and during-training.
Given a trained DNN model, post-training modularization~\cite{pan2020decomposing, pan2022decomposing, imtiaz2023decomposing, cnnsplitter, qi2023reusing, gradsplitter, incite} analyzes how specific model weights contribute to different class predictions by observing the responses of associated hidden units, i.e., neurons in fully connected (FC) layers or channels in convolutional layers.
Subsets of weights that are responsible for particular classes are grouped into individual modules.
However, the inherent inter-connectivity 
of hidden units across layers in a trained model %
yields poor modularity~\cite{csordasneural, sculley2015hidden}, with significant weight overlap between modules~\cite{pan2022decomposing, cnnsplitter, imtiaz2023decomposing}. 
To overcome this issue, more recent work~\cite{mwt} proposed a %
technique that incorporates modular masks into the model's layers to enhance modularity during the training phase.

\looseness-1
Although mask-based modularization offers the promise of improved DNN reuse, it is hindered by three notable limitations.
\circled{1}~%
Achieving modularity within a DNN during training requires an auxiliary mask generator attached to each layer to adjust respective weights' contributions, %
significantly increasing the model complexity and doubling the training time~\cite{mwt}.
\circled{2}~%
Existing masking mechanisms are primarily designed for a DNN's convolutional layers.
However, other types of layers, such as FC layers, are prevalent in DNN architectures and often comprise the majority of a model's parameters (as much as 90\% in VGG)~\cite{basha2020impact}. 
Mask-based modularization approaches end up duplicating the entire set of FC layer weights across modules~\cite{mwt}, 
 obscuring the unique contributions of these weights to module functionalities. %
\circled{3}~%
Despite reduced weight overlaps in convolutional layers, the modules themselves exhibit sub-optimal performance: %
our empirical analysis (see Section~\ref{sec:r-n-r}) suggests that composing these modules into a model yields an  accuracy drop of over 40\%. To mitigate this, additional training of each composed model on the entire sub-task dataset is needed~\cite{mwt}.

\looseness-1
In this paper, we propose \approach, a novel \emph{activation-driven} modular training %
approach that yields fine-grained, accuracy-preserving modules. 
Unlike mask-based approaches that impose modularity \textit{externally} through auxiliary masks, \approach %
fosters modularity \textit{inherently} by directly shaping the activation patterns of DNN layers during training.
Specifically, \approach computes three vital modularization objectives on each layer's activations: \textit{intra-class affinity}, \textit{inter-class dispersion}, and \textit{compactness}. 
In the training phase, \approach aligns the {hidden units} activated across samples within the same class (intra-class affinity), to distinguish the {units} activated for samples from different classes (inter-class dispersion), and to activate minimal numbers of {units} that maintain target functionalities (compactness).

\looseness-1
\approach offers four key benefits.
\circled{1}~\textit{Improved~scalability:} 
By directly regulating the layer outputs without applying masks, \approach avoids introducing extra parameters or modifications to the layers, thus retaining the original model complexity. %
\circled{2}~\textit{Finer~granularity:}
\approach's focus extends %
down to individual neurons within FC layers, thus allowing one to extract only the relevant {hidden units} from the original model to create the requisite modules. %
\circled{3}~\textit{Preserved~module~accuracy:}
\approach's activation-driven approach achieves modularity inherently within the model, facilitating module reuse without the need for additional training to regain accuracy. 
\circled{4}~\textit{Module~replaceability:}
The ability to maintain module accuracy empowers \approach to boost the performance of weak (e.g., overfitting or underfitting) DNN models by replacing their less accurate modules with more accurate ones from stronger models, 
without the need to retrain the entire model.

\looseness-1
We evaluate our work on three representative CNN models (VGG16~\cite{vggmodel}, ResNet18~\cite{resnetmodel} and MobileNet~\cite{mobilenetmodel}) and five widely-used datasets (SVHN~\cite{svhndataset}, CIFAR10~\cite{cifar10dataset}, CIFAR100~\cite{cifar10dataset}, and two subsets of ImageNet~\cite{deng2009imagenet}). 
Our results demonstrate that \approach  promotes modularity inside these models during training and produces fine-grained, compact modules. 
These modules have an average size  and weight overlap as low as 2.1\% and 1.4\%, respectively, compared to the original model, while  maintaining the original classification accuracy.
We also compare {\approach} to  state-of-the-art during-training and post-training modularization approaches, MwT~\cite{mwt}, GradSplitter~\cite{gradsplitter}, and INCITE~\cite{incite}. \approach achieves modules with up to 24x fewer weights and 37x less weight overlap compared to the second-best MwT.
\approach also requires 22\% less training time compared to MwT. 
Finally, in a series of module replacement experiments, \approach demonstrates an average improvement of 12.04\% in accuracy for the target class, with an average increase of 3.52\% in the accuracy of other classes. In contrast, the previous replacement approach, CNNSplitter~\cite{cnnsplitter}, shows a modest~0.91\% improvement for the target classes with the average accuracy increase of~1.99\% for other classes. 

\looseness-1
Our work makes three key contributions.
\circled{1}~A novel activation-driven training method that  yields fine-grained modularity  within DNNs. 
\circled{2}~Integration of three novel objective functions within the training process that generate compact, accuracy-preserving modules. %
\circled{3}~Extensive empirical evaluation,  along with publicly available implementation and artifacts~\cite{website}.

\looseness-1
The paper is organized as follows. 
Section~\ref{sec:key_ideas} discusses the guiding principles behind this work. \approach is detailed in Section~\ref{sec:MODA_approach}. Sections~\ref{sec:evaluation_methodology}, \ref{sec:empirical_results}, and~\ref{sec:ttv} present our evaluation methodology,  results, and validity threats. Related work and conclusions round out the paper.

%% file: sections/2.background.tex
\section{\approach's Guiding Principles}\label{sec:key_ideas}

The limitations of modularization via \emph{external} masks %
motivated us to develop \approach, an approach that supports %
modularity \textit{inherently} within a DNN by directly shaping the activation patterns of DNN layers during training. %
Specifically, \approach's training strategy is built upon two core principles: (1)~improve module specialization by refining activations of hidden units {at a finer-grained level than prior work}, %
and (2)~reduce excessive activations of those units to ensure module compactness. %
We will motivate and briefly introduce both principles next, and then elaborate on them in Section~\ref{sec:MODA_approach}. %

\begin{figure*}[t!]
    \centering
    \setlength{\abovecaptionskip}{3pt}
    \setlength{\belowcaptionskip}{0pt}
    \includegraphics[width=.88\textwidth]{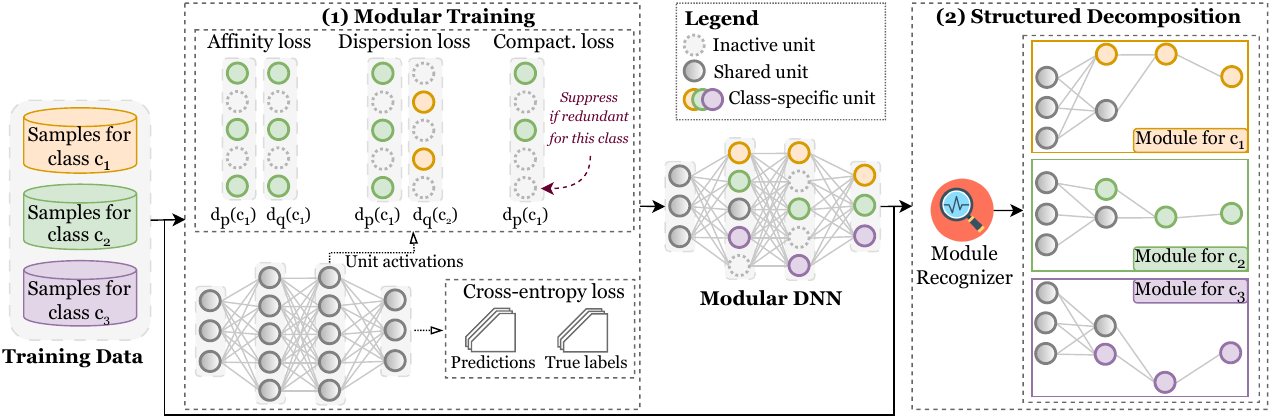}
    \caption{High-level overview of \approach. As depicted in the right-hand segment (2), the number of model's hidden units shared across modules decreases from earlier to later layers.}
    \label{fig:approach_overall_workflow}
    \vspace{-2mm}
\end{figure*}

\looseness-1
\textbf{Inherent Module Specialization} -- 
The key goal of {\approach} is to promote modularity within a multi-layer DNN by cultivating distinct \textit{specialized} sub-networks tailored to different functionalities, i.e., to predict particular classes. %
This functional specialization implies that, ideally, each hidden unit of a layer should activate exclusively in response to input samples from a single class.
Typically, DNNs exhibit a hierarchical structure: earlier (initial) layers share a substantial set of hidden units that capture low-level features (e.g., image edges, textures) across classes~\cite{alexnetmodel}. In later (deeper) layers, neurons are expected to capture more high-level, class-specific concepts (e.g., bird or cat), thereby becoming more specialized and selective in response to input samples from a particular class~\cite{alexnetmodel}.
However, this specialization in DNNs is typically weak because deeper layers may still contain many shared units that are frequently activated for samples from multiple classes~\cite{liu2016towards}.

To remedy this, we propose two novel training objectives that foster the functional specialization of hidden units in each layer by shaping their activation patterns: 
(1) \textit{intra-class affinity} promotes similar subsets of units consistently activated to predict samples within the same class, and
(2) \textit{inter-class dispersion} enforces distinct subsets of units activated to predict samples from different classes.
Throughout  training, these subsets of hidden units will function as sub-networks, which can later be decomposed into  modules. %

\looseness=-1

\looseness-1
Since {\approach} does not introduce additional parameters or modifications to a DNN's layers, the model's original complexity remains unchanged. 
In turn, \approach not only has the potential to enhance the inherent modularity within DNNs, but also to improve the scalability in the training phase.
This allows for finer-grained modularization and broader applicability across varying layer types. For example, the approach that leverages external masks is restricted to channels in convolutional layers, while \approach also works at the level of individual neurons in FC layers.

\textbf{Module Compactness} -- 
In principle, intra-class affinity and inter-class dispersion are sufficient to promote modularity within a DNN. However,  the two objectives %
may  introduce non-essential units.
When such units frequently contribute, even marginally, to class predictions, their  weights become unnecessarily included in the modules.
To address this, we propose the third modularity objective, 
\textit{compactness}, whose direct aim is to ``disable" unnecessary hidden units. %
This is achieved through activation sparsity, induced by gradually reducing non-essential units' activation values toward zero during training (detailed in Section~\ref{sec:MODA_approach}). 

Note that, compactness's goal may appear conceptually similar to model pruning~\cite{ma2019transformed}, which aims to eliminate unnecessary hidden units/weights to reduce the overall model size. However, pruning neurons/weights that appear unimportant at one training epoch permanently removes them from the model, i.e., eliminating their involvement in predicting \textit{any class} in the future. In contrast, \approach's compactness is designed to refine the class-based activation patterns for enhancing DNN modularity. Our intuition behind compactness is to (1) suppress unnecessary activations of neurons for \textit{particular classes} without universally suppressing them for all classes, and (2) allow flexibility for neurons suppressed by compactness in earlier epochs to potentially become active later through the modularity dynamics enforced by affinity and dispersion objectives.

In summary, these three objectives, in tandem, produce modules with minimal yet highly aligned activations, thus enhancing DNN reuse effectiveness, as detailed in the remainder of the paper.

%% file: sections/3.approach.tex
\section{\approach's Approach}\label{sec:MODA_approach}

\looseness-1
Based on the above key ideas, we propose \approach, a novel during-training modularization approach. A high-level view of \approach is shown in Figure~\ref{fig:approach_overall_workflow}. 
Particularly, \approach aims to decompose an $n$-class classification model into $n$ corresponding modules, each containing a subset of weights extracted from the original DNN model that are necessary for predicting the respective class. 
\approach achieves this via two steps:
\circled{1}~\textit{Activation-driven~training:} 
\approach aims to simultaneously reduce the model's prediction errors 
while enhancing its modularity through three principal objectives: intra-class affinity, inter-class dispersion, and compactness. During training, in each layer, activation outputs are gradually refined by promoting specialization where only a limited set of units (e.g., neurons) is involved in making predictions for any class.
\circled{2}~\textit{Structured~decomposition:} 
Once the modular model is trained, \approach identifies the layers' units that are frequently activated while classifying samples of each class. These units, along with their associated weights, are extracted to form distinct modules.
We elaborate each step next, followed by some representative module (re)use scenarios.

\subsection{Activation-Driven Modular Training}\label{sec:modular_training}
This section describes the methodology for training a modular DNN model. 
To facilitate the discussions, we initially focus on fully-connected (FC) layers. We then show how the same principles can be generalized to convolutional layers. 

Formally, an $n$-class DNN model $\mathcal{M}$ is a sequence of  $L$ layers, and we use the following notation:
\[
    \begin{split}
        \mathcal{C} &:= \text{set of classes ($|\mathcal{C}| = n$)} \\ \vspace{-.75mm}
        \mathcal{D} &:= \text{all input samples in a training batch} \\ \vspace{-.75mm}
        \mathcal{D}_c &:= \text{input samples for class } c \in \mathcal{C} \\ \vspace{-.75mm}
        d^c_p &:= p^{\text{th}} \text{ input sample in } \mathcal{D}_c, \forall p \in [|\mathcal{D}_c|] \\ \vspace{-.75mm}
         l &:=  l^{\text{th}}\text{ layer of $\mathcal{M}$}, \forall l \in [1, L] \\ \vspace{-.75mm}
         S^l &:= \text{set of neurons in layer $l$} \\ \vspace{-.75mm}
        \vect{y}^l &:= \text{activation vector for layer $l$}
    \end{split}
\]

Given the activation function $f$, the activation value $y^l_i$ of the $i^{\text{th}}$ neuron $s^l_i$ in layer $l$ is:
 \vspace{-2mm}
\begin{equation}
    ~
    y^l_i = f(\vect{W}^l_i \vect{y}^{l-1} + b^l_i)
    \label{eq:neuron_activation}
\end{equation}

\noindent where $\vect{W}^l_i$ and $b^l_i$ represent $s^l_i$'s  weights and bias, respectively. For notational simplicity, hereafter we omit the layer superscript $l$.

\looseness-1
A neuron $s_i$ is considered as being activated if it has non-zero activation value, i.e., $y_i \neq 0$. Otherwise, $s_i$ is inactive and has no influence on the outputs of the next layer, and consequently does not contribute to final prediction of model $\mathcal{M}$ for the given input $d^c_p$.

Based on this insight, we leverage activation-driven modularity in the training process. 
This involves guiding the model to activate different subsets of neurons $S$ in each layer  for predicting samples of different classes. 
Upon completion of the training phase, we obtain a model $\mathcal{M}$ consisting of $n$ sub-networks for $n$ classes. 

\looseness-1
To this end, we propose a mechanism to dynamically regulate the involvement of all  neurons $S$ in layer $l$ through their activation outputs. 
Our strategy promotes modularity within model $\mathcal{M}$ with three novel objectives: 
(1)~\textit{intra-class affinity}~--~$\mathcal{M}$  activates similar subsets of neurons in $l$ for samples of the same class;
(2) ~\textit{inter-class dispersion}~--~$\mathcal{M}$  activates dissimilar subsets of neurons in $l$ for samples from different classes; and
(3)~\textit{compactness}~--~$\mathcal{M}$  activates a minimal number of neurons in each subset.
As illustrated in Figure~\ref{fig:approach_overall_workflow}, during forward pass, \approach collects the activation vector $\vect{y}$ from layer $l$ corresponding to each input sample. 
\approach subsequently uses these vectors to evaluate the three objectives, in order to guide the optimization in the backward pass toward constructing the activation-driven modularity inside $l$. 
Note that, as with existing work~\cite{pan2020decomposing, pan2022decomposing, qi2023reusing, cnnsplitter, gradsplitter, mwt}, we focus on the ReLU activation function used in hidden layers, as it is the most common activation function for DNNs~\cite{bingham2022discovering}.

\looseness-1
We next detail how the three modularization objectives are computed in the case of FC layers, %
and then show how the same concepts can be applied to convolutional layers.

\textbf{Inter-class Dispersion} -- 
Traditional DNNs use neurons indistinctly across classes, obscuring their specialization~\cite{liu2016towards}. Our goal is to have a clear distinction between subsets of neurons responsible for different classes.
We measure this distinction via the dispersion in activation patterns of neurons responding to input samples.

Given input samples $d^c_p$ from class $c$ and $d^{c'}_q$ from class $c'$, layer $l$ produces activation vectors $(\vect{y})_{d^c_p}$ and $(\vect{y})_{d^{c'}_q}$, respectively. Inter-class dispersion between $d^c_p$ and $d^{c'}_q$ at layer $l$ is computed as:
\vspace{-1mm}
\begin{small}
\begin{equation}
    \textit{dis}(d^{c}_p, d^{c'}_q) = 1 - \textit{sim} \left((\vect{y})_{d^{c}_p} , (\vect{y})_{d^{c'}_q}\right)
    \label{eq:metric_coupling}
\end{equation}
\end{small}

\noindent where $\textit{sim}$ is the similarity measure. 

\looseness-1
In this work, we select cosine similarity since it measures the \textit{directional similarity} between two vectors rather than their magnitudes. 
As with ReLU, the components of activation vector $\vect{y}$ are non-negative, and thus the cosine similarity between two vectors in Equation~\ref{eq:metric_coupling}
is bounded to $[0, 1]$.
A lower similarity score between  $(\vect{y})_{d^{c}_p}$ and $ (\vect{y})_{d^{c'}_q}$ indicates more dispersion. To achieve high inter-class dispersion, the angular distance between two activation vectors belonging to different classes should be as close to orthogonal as possible (i.e., their cosine similarity should approach zero).

In general, given a batch of training samples, every pair of samples  $d^c_p$ and $d^{c'}_q$ from two classes $c$ and $c'$ is selected to measure the dispersion in their resultant activation patterns across layers. To \textit{maximize} dispersion between these patterns, we \textit{minimize} inter-class dispersion loss $\mathcal{L}_{\text{dis}}$ while training the model $\mathcal{M}$, as follows: 

\vspace{-3mm}
\begin{small}
\begin{equation}
    \mathcal{L}_{\text{dis}} = 1 -\frac{1}{|\mathcal{C}| {\cdot} (|\mathcal{C}|-1) / 2} {\sum_{c,c' \in \mathcal{C}}} \left(\frac{\sum_{d^c_p, d^{c'}_q \in \mathcal{D}_c \times \mathcal{D}_{c'}} \textit{dis}(d^{c}_p, d^{c'}_q)}{|\mathcal{D}_c| {\cdot} |\mathcal{D}_{c'}|}\right)
    \label{eq:loss_coupling}
\end{equation}
\end{small}

\looseness-1 \noindent Over a set of training iterations, minimizing inter-class dispersion implies that if one neuron is strongly activated for one class (e.g.,~$c$), its activation intensity for the other class ($c'$) should be reduced toward zero%
, and vice versa.

\textbf{Intra-class Affinity} -- 
This refers to how a model considers a given subset of neurons in $l$ to be responsible for predicting a specific class. The goal of high intra-class affinity is to ensure that the model uses highly similar subsets of neurons when predicting a specific class.
Thus, unlike inter-class dispersion, which assesses the similarity of activation patterns between different classes, intra-class affinity evaluates the similarity of activation patterns within the same class.
Specifically, at  layer $l$, affinity is computed between activation patterns of each pair of samples $d^{c}_p$ and $d^{c}_q$ from the same class $c$ in a training batch:

\vspace{-3mm}
\begin{small}
\begin{equation}
    \textit{aff}(d^{c}_p, d^{c}_q) = \textit{sim} \left((\vect{y})_{d^{c}_p} , (\vect{y})_{d^{c}_q}\right)
    \label{eq:metric_cohesion}
\end{equation}
\end{small}

In turn, \textit{maximizing} intra-class affinity between activation patterns involves \textit{minimizing} the corresponding affinity loss $\mathcal{L}_{\text{aff}}$ during training, which is defined as follows:
\begin{small}
\begin{equation}
    \mathcal{L}_{\text{aff}} = 1 - \frac{1}{\cardinality{\mathcal{C}}} \sum_{c \in \mathcal{C}} \left(\frac{\sum_{d^{c}_p, d^{c}_q \in \mathcal{D}_c} \textit{aff}(d^{c}_p, d^{c}_q)}{|\mathcal{D}_c| {\cdot} \left(|\mathcal{D}_c| - 1\right) / 2}\right)
    \label{eq:loss_cohesion}
\end{equation}
\end{small}

\looseness-1
\textbf{Compactness} -- 
As illustrated in Figure~\ref{fig:approach_overall_workflow}, intra-class affinity loss may enhance neuron involvement, but in the process, could activate additional, non-essential neurons pertaining to a specific class.
On the other hand,  although inter-class dispersion loss reduces shared influence across classes, it may not completely eliminate undesired influences, as neurons may still produce non-zero (even if near-zero) activations.
As a result, these neurons can still marginally affect non-target classes.
To address this, we propose the third objective, compactness, which refines modularity by minimizing the number of neurons involved in predicting a class. 
The compactness loss $\mathcal{L}_{\text{com}}$ is designed based on $l_1$-norm~\cite{ma2019transformed}, and derives the desired properties from it, as discussed below: 

\vspace{-2.5mm}
\begin{small}
\begin{equation}
    \mathcal{L}_{\text{com}} = \frac{1}{\cardinality{\mathcal{C}}} \sum_{c \in \mathcal{C}} \left(\frac{\sum_{d^{c}_p \in \mathcal{D}_c} \lVert{(\vect{y})_{d^c_p}}\rVert_1}{\cardinality{\mathcal{D}_c}}\right)
    \label{eq:loss_compactness}
\end{equation}
\end{small}

\noindent where $\lVert{(\vect{y})_{d^c_p}}\rVert_1$ is the $l_1$-norm of the activation vector $(\vect{y})_{d^c_p}$. 

\looseness-1
During training, $\mathcal{L}_{\text{com}}$ facilitates feature selection, encouraging that only the essential neurons are activated for a class. 
It promotes sparse activation vectors by reducing near-zero activations toward zero, thus resolving the undesired shared influences of neurons encountered in the inter-class dispersion loss.
This differs from previous work~\cite{ma2019transformed}, where the $l_1$-norm was applied to model weights for pruning purposes. In contrast, {\approach} uses it in the compactness objective to refine modularity at the activation level during training.
We will further study the impact of the compactness in Section~\ref{sec:results_rq3}.

In summary, incorporating intra-class affinity, inter-class dispersion, and compactness into our loss function yields a unified loss function $\mathcal{L}$ for training the model $\mathcal{M}$:
\begin{small}
\begin{equation}
    \mathcal{L} = \mathcal{L}_{\text{ce}} + (\alpha \cdot \mathcal{L}_{\text{aff}} + \beta \cdot \mathcal{L}_{\text{dis}} + \gamma \cdot \mathcal{L}_{\text{com}})
    \label{eq:loss_overall}
\end{equation}
\end{small}

\noindent where $\mathcal{L}_{\text{ce}}$ is the standard cross-entropy loss used for evaluating classification errors, and $\alpha$, $\beta$, and $\gamma$ denote weighting factors for the modular losses $\mathcal{L}_{\text{aff}}$, $\mathcal{L}_{\text{dis}}$, and $\mathcal{L}_{\text{com}}$, respectively. 
Each modular loss is averaged across layers before weighting in $\mathcal{L}$.
With mini-batch gradient descent~\cite{eon1998online}, modular training aims to reduce $\mathcal{L}_{\text{ce}}$ to improve the $\mathcal{M}$'s classification accuracy, while decreasing modular losses to enhance $\mathcal{M}$'s modularity.

\textbf{Application to Convolutional Layers} -- Following the same principles, intra-class affinity, inter-class dispersion, and compactness can be applied to convolutional layers.
Unlike an FC layer that produces an activation vector $\vect{y}$, a convolutional layer outputs a 3D activation map $\vect{u} \in \mathbb{R}^{H \times W \times T}$. 
This activation map consists of $T$ channels, each having spatial dimensions of $(H \times W)$. 
Each channel's output $\vect{u}_t$, where $t \in T$, is associated with kernels (i.e., groups of weights) that capture local spatial patterns of the input~\cite{ayinde2019redundant}.
Because of this characteristic, we modularize convolutional layers at channel-level instead of the neuron-level as in FC layers.

To adapt modular objectives to convolutional layers, the modular losses of each layer are estimated from the channel-wise activation vector $\vect{y}$. Specifically, $\vect{y}$ is obtained by performing an averaging operation across the spatial dimensions over the activation map $\vect{u}$:

\vspace{-3.2mm}
\begin{small}
\begin{equation}
    \vect{y} = \frac{1}{H \times W} \sum_{h=1}^{H} \sum_{w=1}^{W} u_{h,w,t}
    \label{eq:cnn_mapping}
\end{equation}
\end{small}

\noindent The resulting scalar $y_t$ for the $t$-th channel is the equivalent of the activation $y_i$ for a neuron in FC layer from Equation~\ref{eq:neuron_activation}. 

Overall, while modular losses apply across all DNN layers, our empirical analysis shows that initial layers still share a majority of neurons across the classes needed to represent low-level features, while distinct class-specific neurons emerge in deeper layers, as depicted in Figure~\ref{fig:approach_overall_workflow}. We discuss this finding in Section~\ref{sec:r-n-r}.

\subsection{Structured Decomposition}
\label{sec:structured_modularization}

\looseness-1
The training phase yields a ``modular model'' $\mathcal{M}$, i.e., a model that is amenable to modularization.
The next step is to decompose $\mathcal{M}$ such that each resulting module retains only a portion of relevant neurons and weights extracted from $\mathcal{M}$.
To decompose $\mathcal{M}$ into $n$ modules corresponding to $n$ classes, \approach measures and selects the highly-utilized neurons in each layer along with their associated weights. %
To preserve individual modules' functionalities, neurons that are frequently activated for two or more classes %
are replicated in the respective modules.
In contrast, neurons that are rarely activated for a class are likely to be redundant and can be excluded from the corresponding module without decreasing its performance.~\footnote{
This decomposition applies to any model, not necessarily one trained with \approach's three objectives.
}
Specifically, the frequency of neuron activation with respect to a particular class $c$ is calculated by the number of times a neuron $s_i$ in layer $l$ is activated in response to input samples from $c$. 
We define $\tau$ as a threshold to determine whether $s_i$ should be included in the module for class $c$ based on the frequency of its activations. 

Formally, the module for class $c$ is defined as:
\begin{small}
\begin{equation}
module(c) = \left\{s_i \mid \forall s_i \in S, \frac{freq(s_i, \mathcal{D}_c)}{|\mathcal{D}_c|} \geq \tau \right\}
\end{equation}
\end{small}

\noindent where \begin{small}$freq(s_i, \mathcal{D}_c)$\end{small} refers to activation frequency of neuron $s_i$ for the set of training samples $\mathcal{D}_c$ of class $c$.

\looseness-1
Since threshold $\tau$ plays a crucial role in the decomposition step, it should be able to balance between module sizes, overlaps, and the potential degradation in classification accuracy.
A higher threshold can generate more lightweight modules but may also omit important neurons and their weights, thus impacting the modules' functionalities.
The choice of  $\tau$ will be discussed further in Section~\ref{sec:results_rq3}.

\subsection{Use Cases for DNN Modularization}
\label{sec:use-case}

\looseness-1
Although the $n$ modules decomposed from $\mathcal{M}$ contain separate groups of neurons and weights, each outputs only a logit (i.e., a scalar) corresponding to its class. 
These modules are not immediately applicable for classification tasks, as a multi-class output vector is needed for determining the most likely class. 
Instead, the modules are intended for  reuse in different scenarios. We detail two such scenarios below; additional scenarios are discussed in Section~\ref{sec:discussion}.

\looseness-1
\textbf{Module Reuse} -- 
To use the model $\mathcal{M}$ for predicting a subset $k$ of its $n$ classes, the corresponding $k$ modules need to be reused in constructing a new model. For example, the two topmost modules in Figure~\ref{fig:approach_overall_workflow} can be composed to create a model  $\mathcal{M'}$ that predicts classes $c_1$ and $c_2$. 
Since these modules are all derived from $\mathcal{M}$, their neurons can be merged layer by layer, with each shared neuron (e.g., the gray neurons in Figure~\ref{fig:approach_overall_workflow}) included only once in the new model.
Since the new model $\mathcal{M'}$ only includes the reused modules' neurons and weights connecting them, the input and output dimensions of the layers are typically smaller than the corresponding layers in the original model.
The last layer of the composed model $\mathcal{M'}$ generates a $k$-dimensional output logit vector indicating the scores of the selected $k$ classes, and can thus compute {$k$-class} probabilities the same way as in $\mathcal{M}$.
Overall, the composed model maintains the same number of layers as $\mathcal{M}$ but with  a fraction of  neuron weights, resulting in reduced memory and computation overhead. 

\looseness-1
Since the modules in this scenario originate from a single  DNN, they are expected to naturally function together when combined for a sub-task.
Therefore, a composed model $\mathcal{M'}$ should {match the accuracy} of the original model $\mathcal{M}$ {without} additional training.
This is one of the key contributions of our work and is distinguished from the state-of-the-art during-training modularization approach MwT~\cite{mwt}, which requires retraining for sub-task reuse.

\begin{figure}[t!]
    \centering
    \setlength{\abovecaptionskip}{3pt}
    \setlength{\belowcaptionskip}{0pt}
    \includegraphics[width=.61\columnwidth]{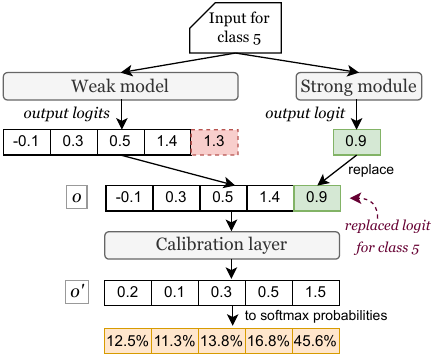}
    \caption{\approach's module replacement strategy}
   \label{fig:replacement_approach}
       \vspace{-6mm}
\end{figure}

\textbf{Module Replacement} --
A common scenario involving integration of modules derived from  different models is substituting a less accurate module $\mathbf{m}^c_w$ associated with a class $c$ and derived from a weak model $\mathcal{M}_{w}$, with a more accurate module $\mathbf{m}^c_s$ for the same class decomposed from a strong model $\mathcal{M}_{s}$~\cite{pan2022decomposing, imtiaz2023decomposing, cnnsplitter, gradsplitter}.
In this context, the model is considered \textit{strong} if it shows higher accuracy %
for class $c$. %
In practice, $\mathcal{M}_{w}$ and $\mathcal{M}_{s}$ may have different architectures.
Thus, integrating module $\mathbf{m}^c_s$ into the base model $\mathcal{M}_w$ on a per-layer basis, as in sub-task reuse, will require additional effort to ensure compatibility.
One approach may be to run $\mathcal{M}_{w}$ and $\mathbf{m}^c_s$ in parallel with a given input, and then replace the output of $\mathcal{M}_w$ for class $c$ with that from $\mathbf{m}^c_s$~\cite{cnnsplitter}.
However, $\mathcal{M}_w$ and $\mathcal{M}_s$ may initially be trained on different datasets, and may potentially share only $c$ as a common class.
Consequently, $\mathcal{M}_{w}$ and $\mathbf{m}^c_s$ may produce outputs (i.e., logits) for their classes in significantly different distributions,
and calculating such outputs' probabilities without any calibration may result in sub-optimal accuracy.

{\approach} allows us to overcome this issue by applying the strategy %
depicted in Figure~\ref{fig:replacement_approach}. 
For each input sample, we denote logit vector $\vect{o}$ obtained from $\mathcal{M}_w$ but with the output logit of target class $c$ replaced by that of $\mathbf{m}^c_s$. 
As discussed above, computing (softmax) probabilities directly on $\vect{o}$ may lead to erroneous predictions due to logit distribution mismatch. %
We thus introduce a \textit{calibration layer}, which is an FC layer that learns to adjust the logits in $\vect{o}$ and produce a calibrated vector $\vect{o'}$ aligned with the true classes.
To achieve this, we freeze all weights in $\mathcal{M}_w$ and $\mathbf{m}^c_s$ and train only the calibration layer using $\mathcal{M}_w$'s training data, including samples from class $c$.
As this layer has a small number of weights, 
it needs only a few training epochs (e.g., 3 to 5) to converge.

%% file: sections/4.evaluation.tex
\section{Evaluation Methodology}\label{sec:evaluation_methodology}

We evaluate \approach by answering three research questions:
\vspace{-1mm}
\begin{itemize}
    \looseness-1
    \item \textbf{RQ1}: How well do the DNN modules decomposed by \approach perform in terms of module reuse and replacement?

    \item \textbf{RQ2}: How effective is \approach in training modular DNNs?

    \item \textbf{RQ3}: How do \approach's underlying design choices impact its performance?
\end{itemize}
\vspace{-1mm}

To answer these questions, we follow the experimental setup used in prior research on DNN modularization~\cite{pan2020decomposing, pan2022decomposing, imtiaz2023decomposing, cnnsplitter, qi2023reusing, gradsplitter, mwt, ren2023deeparc}.

\textbf{Datasets} -- Five datasets are used to evaluate DNN modularization approaches, including Street View House Number (SVHN)~\cite{svhndataset}, CIFAR10~\cite{cifar10dataset},  CIFAR100~\cite{cifar10dataset}, and two variants of ImageNet~\cite{deng2009imagenet}. SVHN and CIFAR10 include 100K and 60K colored $32 \times 32$ images, respectively, classified into 10 classes each. CIFAR100 contains the same number of samples as CIFAR10 but categorized into 100 classes. ImageNet provides 1.3M higher-resolution images, typically at $224 \times 224$, and 1K classes. Due to ImageNet's large scale, we derive two subsets: ImageNet100R, including 133K samples drawn from 100 randomly selected classes, and ImageNet100D, comprising 130K samples from 100 dog-breed classes to evaluate DNN modularization on similar categories. Detailed lists of selected classes are available on our project website~\cite{website}.

\textbf{Models} -- We select three widely-used CNN models VGG16~\cite{vggmodel}, ResNet18~\cite{resnetmodel}, and MobileNet~\cite{mobilenetmodel} with varying sizes and architectures. 
All models contain both convolutional and FC layers in their architecture. 
VGG16 has a sequential architecture built upon stacked layers~\cite{vggmodel}, where outputs of one layer can only flow to the subsequent layer.
ResNet18 relies on blocks of convolutional layers with residual connections, where outputs of one layer can be passed through several layers~\cite{resnetmodel}.
MobileNet's architecture is based on depthwise separable convolutional layers~\cite{mobilenetmodel}, where each output channel of a layer may associate with only one input channel of the immediately preceding layer.
The numbers of parameters for VGG16, ResNet18, and MobileNet on SVHN/CIFAR datasets are around 34.0M, 11.2M, and 3.3M, respectively. 
For ImageNet, VGG16's parameters increase to 134.7M due to larger input dimensions, while the others are mostly unaffected by design~\cite{vggmodel, resnetmodel, mobilenetmodel}.

\textbf{Baselines} -- We compare \approach with a state-of-the-art during-training modularization approach (MwT)~\cite{mwt} and Standard Training (ST), which optimizes the target model using only cross-entropy loss.
For module reuse, we compare the effectiveness of \approach against MwT~\cite{mwt} and two post-training methods, GradSplitter~\cite{gradsplitter} and INCITE~\cite{incite}.
 GradSplitter uses a gradient-based search algorithm to select important convolutional kernels per class and has been shown to outperform prior post-training techniques~\cite{pan2020decomposing,pan2022decomposing,cnnsplitter}. The recently-proposed INCITE estimates neuron clusters' contributions through mutation analysis, and slices modules accordingly. 
Since INCITE supports only stacked-layer DNNs without residual or depthwise blocks~\cite{incite}, we restrict our comparison to VGG16. 
Moreover, both INCITE and GradSplitter become computationally prohibitive at ImageNet scale, where GradSplitter even exceeds our available GPU memory, so we exclude them from our ImageNet experiments.
Finally, we compare \approach's replacement strategy with the one from CNNSplitter~\cite{cnnsplitter} (also employed by GradSplitter) because neither MwT nor INCITE addresses  this scenario.

\looseness-1
\textbf{Hyper-Parameters} -- 
Training for ST, MwT, and \approach is conducted over 200 epochs with a batch size of 128. The mini-batch stochastic gradient descent (SGD) optimizer~\cite{eon1998online} is used with a learning rate of 0.05 and Nesterov's momentum set to 0.9.
For MwT, GradSplitter, and INCITE, we reuse their available implementations and default settings for training and decomposing DNNs~\cite{mwt, gradsplitter, incite}.
For \approach, the weighting factors in the modular losses $\alpha$, $\beta$, and $\gamma$ are set to 1.0, 1.0, and 0.3, respectively. The decomposition threshold $\tau$ is 0.9. We select these hyper-parameters by examining their impact on \approach, as discussed in Section~\ref{sec:results_rq3}.

\looseness-1
\textbf{Evaluation Metrics} -- 
(1)~\textit{Test Accuracy} denotes the top-1 accuracy of the trained model on the test set prior to decomposition into modules.
(2)~\textit{Reuse Accuracy} is the top-1 accuracy of a composed model (i.e., model created by combining modules) on the test set.
(3)~\textit{Module Size} is calculated by dividing the number of weights in a module by the total number of weights in the original model.
Specifically, we quantify weights as individual numerical elements; e.g., a $3 \times 3$ kernel in a convolutional layer is equivalent to 9 weights.
(4)~\textit{Module Overlap} is calculated as the number of weights that are shared between a pair of modules, divided by the total number of weights in the original model. 
(5)~\textit{Composed Model Size} is the number of weights in a composed model divided by the total number of weights in the original model.
(6)~\textit{Composed Model FLOPs} is the total number of floating-point operations (FLOPs) to process an input sample in the composed model divided by that of the original model.

To mitigate the impact of randomness in training DNNs, we set fixed seeds and run  experiments over 5 iterations for SVHN/CIFAR datasets and 3 for ImageNet. We evaluate the final results by averaging across these iterations.
All experiments with \approach and MwT on SVHN/CIFAR are conducted on an Ubuntu 20.04 server with 12 vCPUs, 224 GB of memory, and two NVIDIA  V100 GPUs  with 16 GB of VRAM each. All other experiments are conducted on a similar machine, but with 64 vCPUs, 526 GB of memory, and an NVIDIA GH200 with 96 GB of VRAM.

%% file: sections/5.results.tex
\section{Empirical Results}\label{sec:empirical_results}

\subsection{RQ1 -- Module Reuse and Replacement}
\label{sec:r-n-r}
The goal of this research question is to evaluate the support for reuse and replacement of decomposed modules produced by {\approach}, 
which reflects the effectiveness of its overall modularization process.

\input{figures/rq2/module_reuse_figure}

\textbf{Module Reuse} --
To assess the quality of the $n$ modules derived from modular model $\mathcal{M}$, they are selectively assembled to compose a new model tailored to a specific sub-task, i.e., classifying a subset of $n$ classes (recall Section~\ref{sec:use-case}).
Specifically, each $k$-class sub-task type involves choosing $k$ classes from a set of $n$ classes, resulting in a total of $n \choose k$ possible $k$-class sub-tasks.
For SVHN and CIFAR10 datasets, each with 10 classes, we define 9 sub-task types ranging from 2 to 10 classes, yielding 1,013 sub-tasks in total.
Similarly, for the CIFAR100, ImageNet100R, and ImageNet100D datasets, we define 99 sub-task types ranging from 2 to 100 classes.
Due to the extremely large number of possible combinations for each sub-task type, we randomly sample a representative subset of all possible tasks with 95\% confidence level and 8\% margin of error, resulting in 14,697 sub-tasks. 

\looseness-1
We evaluate reuse accuracy for each sub-task type based on the average classification accuracy of the composed models, as defined in Section~\ref{sec:evaluation_methodology}.
The quality of the modules from {\approach}, MwT~\cite{mwt}, GradSplitter~\cite{gradsplitter}, and INCITE~\cite{incite} correlates with how closely the reuse accuracy of the composed models built using these modules matches that of the \textit{full DNN} trained with Standard Training (ST).\footnote{The post-training baselines, GradSplitter and INCITE, extract modules from this ST's pre-trained model.}
Critically, {\approach} {does not involve any retraining or fine-tuning} after composing models, while MwT necessitates retraining of the entire composed models on  sub-task datasets to regain accuracy~\cite{mwt}.
Although retraining composed models may in practice improve reuse accuracy for the targeted sub-tasks, it conceals the actual quality of the underlying reused modules.
To provide a directly comparable evaluation, we compare {\approach} to the non-retraining variant of MwT. %
For completeness, results for MwT with retraining are also provided in our online appendix~\cite{website}.

\input{figures/rq3/replacement_for_overfitting_model}

\looseness-1
We evaluate ST, \approach, MwT, GradSplitter, and INCITE on three CNN models and five datasets.
Due to space constraints, we only present  results for the experiments conducted with VGG16 on CIFAR10 and CIFAR100. 
The remaining combinations of models and datasets follow a similar trend, detailed in~\cite{website}.
As depicted in Figures~\ref{fig:rq2_composed_model_accuracy_cifar10} and ~\ref{fig:rq2_composed_model_accuracy_cifar100}, \approach's composed models achieve comparable accuracy to ST's models, and significantly outperform the competing modularization methods.
Across all sub-tasks in VGG16-CIFAR10 (Figure~\ref{fig:rq2_composed_model_accuracy_cifar10}), the average reuse accuracies achieved by ST, \approach, MwT, GradSplitter, and INCITE are 95.2\%, 94.5\%, 47.9\%, 94.7\%, 73.8\%, respectively.
However, when scaling to the more complex CIFAR100 dataset (Figure~\ref{fig:rq2_composed_model_accuracy_cifar100}), only \approach's accuracy remains on par with ST's (both at 78.2\%), while GradSplitter drops to 47.3\%, followed by INCITE at 39\% and MwT at 6.8\%.
The same pattern holds for ResNet18 and MobileNet models.
We further evaluated on the two ImageNet datasets, focusing solely on \approach and MwT for reasons discussed in Section~\ref{sec:evaluation_methodology}. \approach's reuse accuracies remain comparable to ST, with the gap of under 1\%, whereas MwT's gap is above 71\%.

Achieving compact modules is another goal of \approach. 
Considering that lightweight modules are only valuable if their performance is comparable to ST, we evaluate the modules' sizes alongside their reuse accuracies.
In VGG16-CIFAR10, the average module sizes (defined in Section~\ref{sec:evaluation_methodology}) are 2.7\%, 52.2\%, 58.7\%, and 52.0\% for \approach, MwT, GradSplitter and INCITE, respectively.
The average module overlaps yielded by the four techniques are 1.4\%, 50.5\%, 55.3\%, 51.0\%, respectively.
Furthermore, only \approach and GradSplitter retain their reuse accuracies comparable to ST, as shown in Figure~\ref{fig:rq2_composed_model_accuracy_cifar10}.
In the case of VGG16-CIFAR100, \approach's average module size and overlap increase to 5.9\% and 3.4\%, respectively, but are still significantly lower than those of the second-best method, MwT (51.8\% and 51.0\%); only \approach preserves accuracy at this scale (Figure~\ref{fig:rq2_composed_model_accuracy_cifar100}).
Overall, across all of the model-dataset pairs, \approach reduces module size and overlap by up to 24x and 37x relative to its closest competitor. In some cases, e.g., ResNet18–CIFAR100, MwT yields smaller module sizes (3\% vs. \approach's 23\%) but suffers from significantly lower reuse accuracy (4.6\% vs. 78.7\%).

We also conducted an empirical study to examine overlap between \approach's modules across different layers.
Our results show that there is substantial weight overlap between modules in initial layers; this overlap decreases significantly as we move toward deeper layers. 
For instance, in the VGG16-CIFAR10 model, the average weight overlaps over total weight of a layer between module-pairs are more than 99\% for the first three convolutional layers but less than 0.15\% in the last three convolutional layers.
This indicates that modules share a substantial set of hidden units in the initial layers, which is typically known to capture low-level features (e.g., edges and corners in input images)~\cite{liu2016towards, alexnetmodel}. In contrast, each module has a more distinct, class-specific set of hidden units in the higher layers to represent more abstract, class-specific concepts (e.g., bird or cat).

\looseness-1
Considering the reuse scenarios, we present representative composed model sizes and FLOPs for \approach, MwT, GradSplitter, and INCITE in Figures~\ref{fig:rq2_composed_model_size_cifar10} and ~\ref{fig:rq2_composed_model_size_cifar100}. 
\approach and MwT  compose modules into a single model, guaranteeing  the resulting model's size and FLOPs do not exceed the original's ($\leq$100\%). By contrast, GradSplitter and INCITE  run multiple modules independently (and only merge the outputs) for each input~\cite{mwt, gradsplitter, incite}. This means that their combined size and FLOPs are the sum of these modules and can surpass 100\%. 
In  VGG16-CIFAR10, models composed by \approach exhibit 70.7\%, 93.9\%, and 93.1\% fewer weights than those from MwT, GradSplitter, and INCITE, respectively, across sub-tasks. In VGG16–CIFAR100, the corresponding reductions are 12.3\%, 96.9\%, and 97.1\%. As another example (elided for space), for  ResNet18-CIFAR10, \approach yields 46.5\% fewer weights than the second-best MwT~\cite{website}.

\looseness-1
In terms of FLOPs, in the VGG16-CIFAR10 model, {\approach} yields average reductions of 5.9\%, 80.4\%, and 78.9\% relative to MwT, GradSplitter, and INCITE, respectively. In VGG16–CIFAR100, \approach yields 14.5\% more FLOPs than MwT, while reducing FLOPs by 94.4\% and 95.2\% compared to GradSplitter and INCITE, respectively. For the ResNet18-CIFAR10 model (elided for space), the reductions are  1.31\% for MwT and 78.7\% for GradSplitter; INCITE does not support ResNet18. 
In Figure~\ref{fig:rq2_composed_model_size_cifar10}, the sharp decrease in \approach's model sizes %
is not reflected in the required FLOPs because most of the computation is in convolutional~layers~\cite{li2016pruning}.
For example, our analysis shows that convolutional layers occupy 43.8\% of weights but account for 94.3\% of FLOPs in the VGG16-CIFAR10 model, with the remainder attributed to the FC layers.

\input{figures/rq1/modular_training_mod_metrics_merged}

\looseness-1
\textbf{Module Replacement} --
We compare replacement strategies in {\approach} and CNNSplitter~\cite{cnnsplitter}, specifically focusing on substituting a module from a weak model $\mathcal{M}_{w}$ having low accuracy with a higher accuracy module $\mathbf{m}^c_s$ for the same class $c$ decomposed from a strong model $\mathcal{M}_{s}$.
Since the replacement strategies of {\approach} and CNNSplitter are entirely independent of their training and decomposition steps, we use the same set of modules generated by our modularization process as inputs for both replacement approaches.
Specifically, we train VGG16~\cite{vggmodel} and ResNet18~\cite{resnetmodel} and use them as strong models.
To build a weak model $\mathcal{M}_{w}$, we use LeNet5~\cite{lenet5}, which comprises a stack of three convolutional 
and two FC layers.
We conduct the experiments on two common types of weak DNN models: overfitted and underfitted~\cite{ma2018mode}.
To obtain overfitted models, which have perfect accuracy on the training set and low accuracy on the test set, we randomly sample 10\% of training data and disable data augmentation~\cite{shorten2019survey}, dropout~\cite{srivastava2014dropout}, and weight decay~\cite{krogh1991simple}, as done in prior work~\cite{cnnsplitter}.   
To obtain underfitted models, which have low accuracy on both sets, we train LeNet5 with only 5\% of {the standard 200 epochs} and evaluate its performance~\cite{cnnsplitter}.

We conduct the experiments on SVHN and CIFAR10 datasets, each containing 10 classes, denoted as $C = \{c_1, c_2, \ldots, c_{10}\}$.
Each dataset is split into two subsets of samples for training $\mathcal{M}_{s}$ and $\mathcal{M}_{w}$.
Subset of samples from classes $C_s = \{c_1, c_2, \cdots, c_{5}\}$ are used to train strong model $\mathcal{M}_{s}$, while samples from  $C_w = \{c_6, c_7, \cdots, c_{10}, c_w\}$, where $c_w \in C_s$, are used to train weak model $\mathcal{M}_{w}$. 
In this setup, $\mathcal{M}_{s}$ and $\mathcal{M}_{w}$ share the common class $c_w$, which is the target class for improvement. %
We repeat choosing each $c_w$ in $C_s$, which results in five different datasets for training $\mathcal{M}_{w}$.

\looseness-1
Table~\ref{tab:replacement_accuracy} reports the accuracy differences between pre- and post-replacement for the target class (TC) and the non-target classes (non-TC) in  the model overfitting and  underfitting scenarios. 
\approach outperforms CNNSplitter in accuracy improvement of target classes in $39/40$ cases. 
In particular, \approach boosts a TC's accuracy for overfitted models in $20/20$ cases by an average of 10.11\%, and for underfitted models in $18/20$ cases by 13.97\%.
In the two cases where \approach shows a reduction in accuracy, the initial accuracy of TC in weak model $\mathcal{M}_{w}$ is already high (95\%) and replacement nearly matches it (94.47\% and 94.43\%).
Meanwhile, CNNSplitter improves TC's accuracy in only 10/20 cases for overfitted models (0.9\% average improvement) and 6/20 cases for underfitted models (0.31\% average improvement).

\looseness-1
We also study how replacing the TC in $\mathcal{M}_{w}$ affects the accuracy of non-TCs. 
Table~\ref{tab:replacement_accuracy} suggests that both \approach and CNNSplitter exhibit negligible changes (0.28\% and 1.42\%, respectively) in the average accuracy of overfitted models classifying other classes. 
The accuracy differences are more noticeable for underfitted models (6.76\% and 2.56\%, respectively).
In both scenarios, the primary reason for the improved non-TCs accuracy is from the samples previously misclassified as TC, correctly reclassified as non-TCs after the replacement.

\subsection{RQ2 -- Modular Training}\label{sec:results_rq2}

\looseness-1
Modular training is the most resource-intensive phase of the modularization process. More importantly, the performance of modular models produced by this phase directly impacts subsequent steps, i.e., module decomposition and composition.
Therefore, this research question aims to evaluate the effectiveness of {\approach} in training DNN models by measuring their test accuracy and training time.
Table~\ref{tab:training_accuracy} reports the average test accuracy along with standard deviations for ST, MwT, and \approach for three CNN models and five datasets.
Overall, the average accuracy of ST, MwT, and \approach is 83.9\%, 63.0\%, 83.7\%, respectively.

\textbf{Test Accuracy} --
The accuracy of models trained by \approach is comparable to ST across different datasets and models: it is lower by 0.2\% on average, and actually surpasses ST in certain cases (e.g., VGG16-CIFAR100, ResNet18-ImageNet100D). On the other hand, MwT's accuracy decreases significantly in a number of cases. For example, when applied on the \emph{large} CIFAR100, ImageNet100D, and ImageNet100R datasets, MwT's accuracy drops by nearly 23\% on average, 
and when used with the relatively \emph{small} MobileNet model its accuracy loss is about 50\%.
We hypothesize that the reason for the former is that MwT is unable to cope with CIFAR100 or ImageNet-level datasets' size and complexity. In the latter case,  MobileNet's  small size and resulting limited weight space hinder MwT in guiding MobileNet to learn sufficient features while still ensuring that MwT's masks can regulate the modularity within the model.
It is also worth noting MobileNet is built on a different architecture than the other two models (recall Section~\ref{sec:evaluation_methodology}).
Meanwhile, \approach directly promotes the inherent modularity by regulating the activation outputs of each layer, thereby improving its generalizability across different architectures and datasets of varying complexity.

\textbf{Training Time} -- %
In comparison to ST, the average runtime overhead of \approach and MwT is 33\% and 79\%, respectively.
Overall, \approach demonstrates faster training time compared to MwT across all cases, with an average reduction of 22\%, even though MwT only supports modularity within convolutional layers. 
The difference in efficiency between two approaches primarily comes from the additional weights incurred by the modular masks in MwT, which need to be optimized alongside  the original model's weights.
On the other hand, although {\approach} requires more training time than ST, it offers long-term advantages by allowing the reuse and composition of new DNNs to adapt to new requirements without retraining.

\input{figures/rq2/impact_of_compactness/compactness}

\subsection{RQ3 -- Impact of Design Choices}\label{sec:results_rq3}

\looseness-1 This research question studies the impact of \approach's three  modularization objectives and the hyper-parameters used in \approach.

\textbf{Modularization Objectives} --
Section~\ref{sec:r-n-r} discussed how {\approach} leverages its loss functions -- intra-class affinity ($\mathcal{L}_{\text{aff}}$), inter-class dispersion ($\mathcal{L}_{\text{dis}}$), and compactness ($\mathcal{L}_{\text{com}}$) -- to produce fine-grained, accuracy-preserving, reusable modules.
Here, we investigate how these  objectives  affect the quality of the modules generated by {\approach}.
Specifically, since affinity and dispersion are designed to operate in tandem (recall Section~\ref{sec:key_ideas}),  we study two configurations of \approach: (1)~{\approach} with intra-class affinity  and inter-class dispersion (but with  compactness removed), referred to as $\text{\approach}^{-}$, and (2) {\approach} with all three objectives included.
Following the modularization procedure detailed in Section~\ref{sec:MODA_approach}, we evaluate the resultant modules on three metrics as defined in Section~\ref{sec:evaluation_methodology}: (1)~reuse accuracy, (2)~module size, and (3)~module overlap.
We calculate the three metrics after the modularization across all employed models and datasets, for both configurations of \approach. 
For reuse accuracy, we compute the average accuracy of composed models across all sub-task types.

\looseness-1
Table~\ref{tab:impact_of_compactness} presents the results we obtained. %
Overall, $\text{\approach}^-$ produces modules with only 33.4\% and 30.2\% in terms of average module size and overlap, while their reuse accuracy matches the standard model. 
Additionally, we observe that incorporating $\mathcal{L}_{\text{com}}$ in {\approach} leads to much more compact modules, with 17.8\% fewer total weights and 16.5\% fewer weight overlap as compared to {$\text{\approach}^-$}, while retaining comparable reuse accuracy (0.2\% average accuracy loss).
In terms of the absolute number of reduced modules' weights, the differences between \approach and $\text{\approach}^-$ across the three models %
are significant. 
For example, in the case of the CIFAR10 dataset, \approach yields 288K, 423K, and 75K fewer weights for each module from VGG16, ResNet18, and MobileNet, respectively.

\textbf{Impact of Hyper-Parameters} --
Finally, we investigate the influence of hyper-parameters on the modular training process and the impact of the threshold $\tau$ on module decomposition. Recalling the unified loss function from Section~\ref{sec:modular_training}, we study the influence of weighting factors $\alpha$, $\beta$, and $\gamma$ pertaining to intra-class affinity, inter-class dispersion, and compactness, respectively. Specifically, we start with the default configuration $\{\alpha=1.0, \beta=1.0, \gamma=0.3\}$ and change a single hyper-parameter at a time to compute the training losses and test accuracy. Due to space limitations, we only summarize our findings. All results are available in our appendix~\cite{website}. %

\begin{figure}[b!]
    \vspace{-3.5mm}
    \centering
    \setlength{\abovecaptionskip}{3pt}
    \setlength{\belowcaptionskip}{0pt}
     \includegraphics[width=.7\columnwidth]{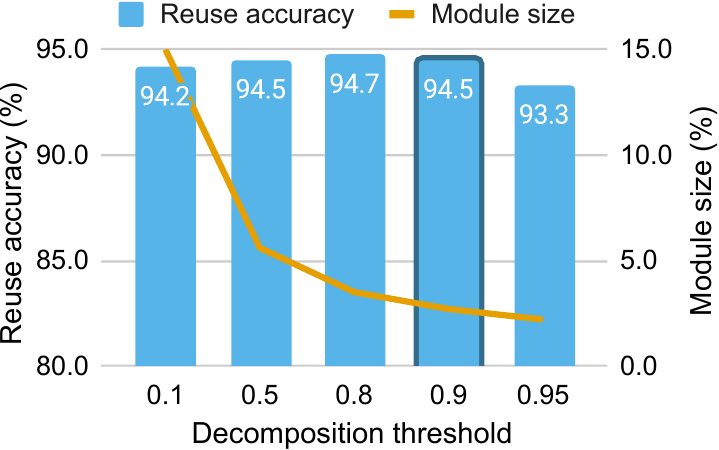}
     \caption{(RQ3) Impact of threshold $\tau$ on modularization}
     \label{fig:rq4_modularizing_threshold}
\end{figure}

We observed that, for higher values of $\alpha$, intra-class affinity loss reduces and test accuracy goes up modestly. However, larger $\alpha$ also results in higher dispersion and compactness losses.
Higher $\beta$ %
 values contribute to lower inter-class dispersion loss while increasing affinity loss. Higher $\beta$ values also result in lower compactness loss as well as lower test accuracy.
Finally, changing $\gamma$ values %
yielded negligible differences for affinity and dispersion losses. However, higher $\gamma$ values can negatively impact test accuracy.
In summary, the use of extreme values for both the $\alpha$ and $\beta$ hyper-parameters substantially affects other losses. Additionally, we found that the changing values for $\gamma$ have negligible effect on affinity and dispersion losses but influence test accuracy. 
Based on this observation, the proposed default hyper-parameters are employed to achieve a balance among the modular losses and test accuracy.

\looseness-1
We evaluate the impact of decomposition threshold $\tau$ (recall Section~\ref{sec:structured_modularization}) on reuse accuracy and module size by varying it between 0.1, 0.5, 0.8, 0.9, and 0.95. 
As shown in Figure~\ref{fig:rq4_modularizing_threshold}, setting $\tau$ to 0.8 yielded the highest accuracy, but 0.95 resulted in the smallest module size. 
To balance accuracy and size, we used 0.9 as default $\tau$ value.

%% file: figures/rq2/module_reuse_figure.tex
\begin{figure*}[t]
    \vspace{-2mm}
    \centering
    
    \begin{subfigure}[b]{0.47\linewidth}
        \centering
        \includegraphics[width=\linewidth]{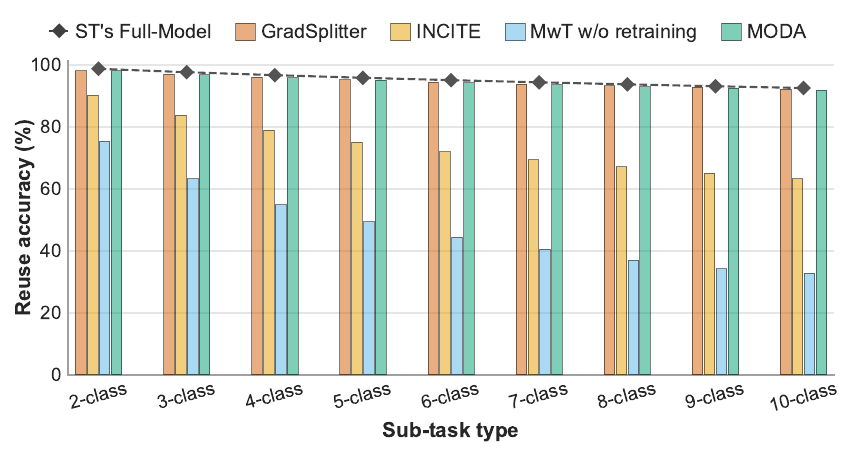}
        \vspace{-7mm}
        \caption{Reuse accuracy of VGG16–CIFAR10}
        \label{fig:rq2_composed_model_accuracy_cifar10}
    \end{subfigure}
    \hfill
    \begin{subfigure}[b]{0.475\linewidth}
        \centering
        \includegraphics[width=\linewidth]{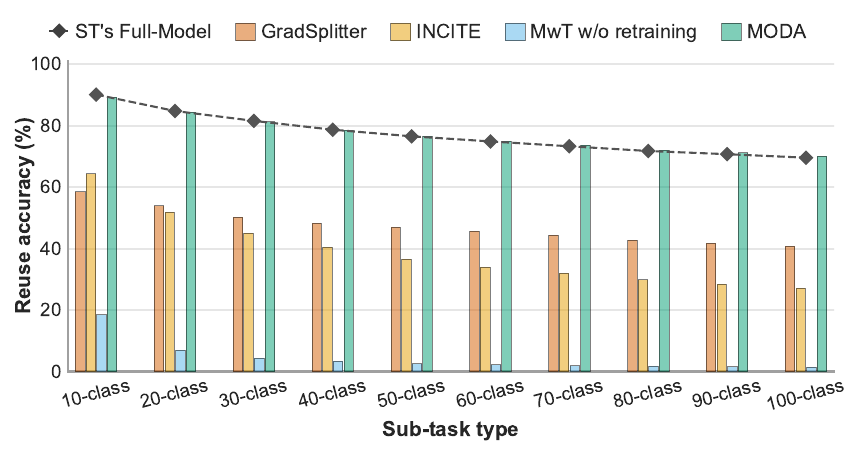}
        \vspace{-7mm}
        \caption{Reuse accuracy of VGG16–CIFAR100}
        \label{fig:rq2_composed_model_accuracy_cifar100}
    \end{subfigure}

    \vspace{-1mm}\par\bigskip

    \begin{subfigure}[b]{0.47\linewidth}
        \centering
        \includegraphics[width=\linewidth]{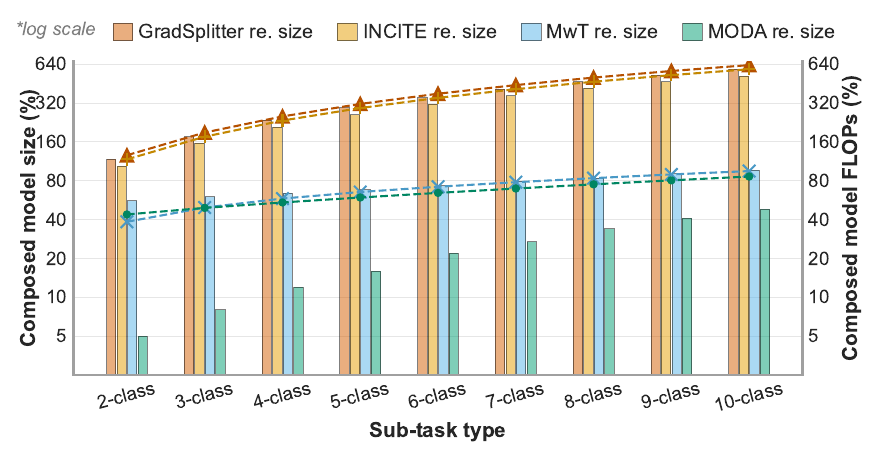}
        \vspace{-7mm}
        \caption{Composed model sizes (bars) and FLOPs (curves) of VGG16–CIFAR10}
        \label{fig:rq2_composed_model_size_cifar10}
    \end{subfigure}
    \hfill
    \begin{subfigure}[b]{0.475\linewidth}
        \centering
        \includegraphics[width=\linewidth]{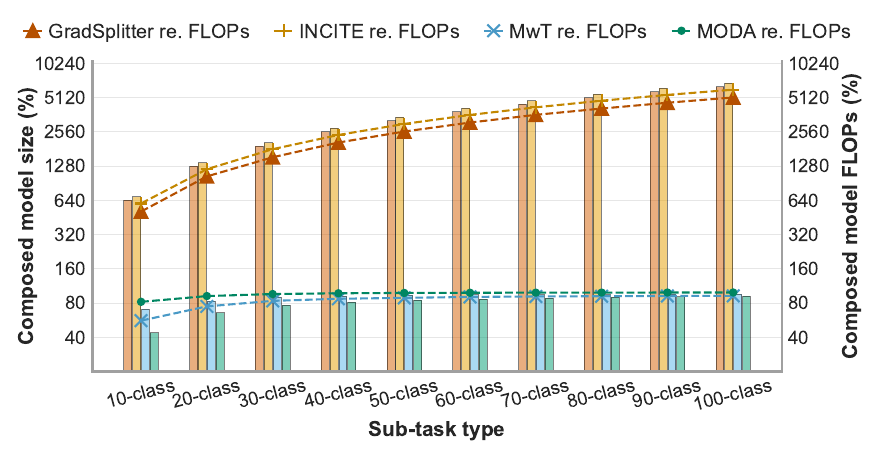}
        \vspace{-7mm}
        \caption{Composed model sizes (bars) and FLOPs (curves) of VGG16–CIFAR100}
        \label{fig:rq2_composed_model_size_cifar100}
    \end{subfigure}

    \vspace{-3mm}
    \caption{(RQ1) Measurements of module reuse across different sub-task types for the VGG16 model on CIFAR datasets}
    \label{fig:rq2_composed_model}
    \vspace{-2mm}
\end{figure*}

%% file: figures/rq3/replacement_for_overfitting_model.tex
\begin{table*}[t]
\centering
\setlength{\abovecaptionskip}{3pt}
\setlength{\belowcaptionskip}{0pt}
\caption{(RQ1) Module replacement comparison of \approach and \emph{CNNS}plitter for improving  the accuracy of \emph{ove}r/\emph{und}erfitted models}
\label{tab:replacement_accuracy}
\renewcommand{\arraystretch}{1.05}
\resizebox{\textwidth}{!}{%
\begin{tabular}{l|l|ccccc|ccccc}
\toprule
\multirow{2}{*}{\textbf{Weak Model}} & \multirow{2}{*}{\textbf{Appr.}} & \multicolumn{5}{c|}{\textbf{Accuracy Difference in TC $|$ non-TC (\%) for CIFAR10}} & \multicolumn{5}{c}{\textbf{Accuracy Difference in TC $|$ non-TC (\%) for SVHN}} \\
\cmidrule{3-12}
\textbf{w/ Strong Model} & & \textbf{airplane} & \textbf{automobile} & \textbf{bird} & \textbf{cat} & \textbf{deer} & \textbf{Digit 0} & \textbf{Digit 1} & \textbf{Digit 2} & \textbf{Digit 3} & \textbf{Digit 4} \\ 
\midrule
LeNet5 (ove) & CNNS & 3.10$|$0.38 & -3.40$|$2.64 & -2.10$|$1.56 & -0.90$|$2.38 & 0.60$|$1.84 & -1.78$|$0.88 & -1.75$|$0.82 & -0.58$|$0.91 & -1.39$|$1.29 & -0.44$|$0.68 \\
w/ VGG16 & \approach & \textbf{8.50$|$-0.22} & \textbf{17.00$|$0.40} & \textbf{12.50$|$-0.14} & \textbf{5.90$|$0.48} & \textbf{13.80$|$0.42} & \textbf{2.98$|$-0.29} & \textbf{1.66$|$0.33} & \textbf{3.04$|$-0.66} & \textbf{3.88$|$-0.69} & \textbf{5.43$|$-0.14} \\\midrule
LeNet5 (ove) & CNNS & 5.20$|$0.08 & ~4.60$|$0.70 & -1.20$|$1.58 & 0.20$|$1.80 & 2.90$|$1.22 & 2.12$|$0.28 & 0.39$|$0.56 & 0.87$|$0.70 & -0.32$|$1.06 & 1.27$|$0.46 \\
w/ ResNet18& \approach & \textbf{7.30$|$-0.40} & \textbf{10.10$|$0.34} & \textbf{12.80$|$0.58} & \textbf{5.20$|$1.02} & \textbf{8.00$|$0.28} & \textbf{3.84$|$-0.13} & \textbf{1.33$|$0.17} & \textbf{2.82$|$-0.15} & \textbf{2.43$|$-0.03} & \textbf{5.70$|$0.27} \\
\midrule
\midrule
LeNet5 (und) & CNNS & 10.00$|$-0.78 & -8.20$|$3.26 & -6.00$|$5.04 & -5.90$|$5.10 & -4.60$|$3.52 & -4.01$|$1.20 & -2.85$|$1.61 & -4.03$|$2.09 & -8.54$|$2.75 & -1.23$|$0.94 \\
w/ VGG16 & \approach & \textbf{10.30$|$3.50} & \textbf{17.50$|$6.76} & \textbf{16.70$|$6.82} & \textbf{10.20$|$5.64} & \textbf{24.00$|$10.22} & \textbf{4.25$|-$0.19} & \textbf{-0.63$|$1.51} & \textbf{4.00$|$1.15} & \textbf{1.46$|$1.42} & \textbf{3.96$|$0.58} \\
\midrule
LeNet5 (und) & CNNS & \textbf{16.60$|$-2.86} & 4.20$|$1.00 & -3.00$|$5.04 & -1.90$|$3.54 & 1.90$|$2.74 & 3.79$|$0.17 & -0.75$|$1.07 & -0.56$|$1.92 & -3.44$|$1.46 & 0.95$|$0.82 \\
w/ ResNet18& \approach & 9.30$|$3.40 & \textbf{9.60$|$6.00} & \textbf{16.20$|$8.18} & \textbf{7.60$|$7.42} & \textbf{18.30$|$9.64} & \textbf{6.14$|$0.52} & \textbf{-0.67$|$1.65} & \textbf{4.57$|$2.14} & \textbf{0.10$|$2.48} & \textbf{4.60$|$0.57} \\
\bottomrule
\end{tabular}}
\vspace{-3mm}
\end{table*}

%% file: figures/rq1/modular_training_mod_metrics_merged.tex
\begin{table*}[t!]
\centering
\setlength{\abovecaptionskip}{3pt}
\setlength{\belowcaptionskip}{0pt}
\caption{{(RQ2) Comparison of models trained with {\approach} and MwT relative to Standard Training (ST), prior to decomposition. Rows marked with ($\dagger$) indicate experiments run on the NVIDIA GH200 GPU, while all others used the V100.}}
\label{tab:training_accuracy}
\renewcommand{\arraystretch}{1.05}
\resizebox{.7\textwidth}{!}{%
\begin{tabular}{l|l|ccc|ccc}
\toprule
\multirow{2}{*}{\textbf{Model}} & \multirow{2}{*}{\textbf{Dataset}} &
\multicolumn{3}{c|}{\textbf{Test Accuracy (\%)}} &
\multicolumn{3}{c}{\textbf{Training Time (hours)}} \\
\cmidrule{3-8}
& & \textbf{ST} & \textbf{MwT} & \textbf{\approach} & \textbf{ST} & \textbf{MwT} & \textbf{\approach} \\
\midrule
\multirow{5}{*}{VGG16}
 & SVHN             & 95.85 $\pm$ 0.06 & 95.06 $\pm$ 0.11 & \textbf{95.78 $\pm$ 0.04} & 1.28 & 2.10 & \textbf{1.76} \\
 & CIFAR10          & 92.62 $\pm$ 0.43 & 90.55 $\pm$ 0.13 & \textbf{91.86 $\pm$ 0.23} & 0.81 & 1.38 & \textbf{1.13} \\
 & CIFAR100         & 69.46 $\pm$ 0.31 & 62.48 $\pm$ 0.16 & \textbf{70.11 $\pm$ 0.17} & 0.82 & 1.39 & \textbf{1.14} \\
 & ImageNet100R$^\dagger$ & 85.11 $\pm$ 0.27 & 71.41 $\pm$ 3.44 & \textbf{85.63 $\pm$ 0.27} & 7.86 & 9.96 & \textbf{9.34} \\
 & ImageNet100D$^\dagger$ & 80.31 $\pm$ 0.49 & 67.46 $\pm$ 4.45 & \textbf{80.45 $\pm$ 0.70} & 7.78 & 9.76 & \textbf{9.14} \\
\midrule
\multirow{5}{*}{ResNet18}
 & SVHN             & 96.12 $\pm$ 0.04 & 95.71 $\pm$ 0.05 & \textbf{95.75 $\pm$ 0.05} & 1.90 & 3.35 & \textbf{2.48} \\
 & CIFAR10          & 93.37 $\pm$ 0.07 & \textbf{91.04 $\pm$ 0.03} & 90.94 $\pm$ 0.07 & 1.23 & 2.18 & \textbf{1.61} \\
 & CIFAR100         & 72.82 $\pm$ 0.23 & 60.02 $\pm$ 0.66 & \textbf{70.79 $\pm$ 0.16} & 1.23 & 2.19 & \textbf{1.62} \\
 & ImageNet100R$^\dagger$ & 85.57 $\pm$ 0.02 & 77.45 $\pm$ 0.47 & \textbf{85.25 $\pm$ 0.46} & 15.33 & 20.32 & \textbf{18.15} \\
 & ImageNet100D$^\dagger$ & 78.15 $\pm$ 0.22 & 72.73 $\pm$ 1.65 & \textbf{78.49 $\pm$ 0.39} & 14.92 & 19.84 & \textbf{17.75} \\
\midrule
\multirow{5}{*}{MobileNet}
 & SVHN             & 95.04 $\pm$ 0.00 & 32.87 $\pm$ 35.12 & \textbf{95.00 $\pm$ 0.54} & 1.10 & 2.84 & \textbf{1.61} \\
 & CIFAR10          & 89.42 $\pm$ 0.25 & 53.55 $\pm$ 5.27  & \textbf{89.34 $\pm$ 0.07} & 0.66 & 1.83 & \textbf{1.03} \\
 & CIFAR100         & 62.10 $\pm$ 0.33 & 31.51 $\pm$ 2.42  & \textbf{62.58 $\pm$ 0.21} & 0.68 & 1.93 & \textbf{1.05} \\
 & ImageNet100R$^\dagger$ & 85.31 $\pm$ 0.02 & 23.51 $\pm$ 38.98 & \textbf{85.13 $\pm$ 0.01} & 7.95 & 12.43 & \textbf{10.27} \\
 & ImageNet100D$^\dagger$ & 77.54 $\pm$ 0.38 & 19.77 $\pm$ 32.50 & \textbf{77.82 $\pm$ 0.69} & 7.69 & 12.13 & \textbf{10.06} \\
\bottomrule
\end{tabular}}
\vspace{-5pt}
\end{table*}

%% file: figures/rq2/impact_of_compactness/compactness.tex
\begin{table*}[]
\centering
\setlength{\abovecaptionskip}{3pt}
\setlength{\belowcaptionskip}{0pt}
\caption{{(RQ3) Impact of three modular objectives on \approach}}
\label{tab:impact_of_compactness}
\renewcommand{\arraystretch}{1.05}
\resizebox{.73\textwidth}{!}{
\begin{tabular}{l|l|cc|ccc|ccc}
\toprule
\multirow{2}{*}{\textbf{Model}} & \multirow{2}{*}{\textbf{Dataset}} & \multicolumn{2}{c|}{\textbf{Reuse Accuracy}} & \multicolumn{3}{c|}{\textbf{Module Size}} & \multicolumn{3}{c}{\textbf{Module Overlap}} \\
\cmidrule{3-10}
& & \textbf{\approach} & $\text{\approach}^{-}$ & \textbf{\approach} & $\text{\approach}^{-}$ & \textbf{Reduction} & \textbf{\approach} & $\text{\approach}^{-}$ & \textbf{Reduction} \\
\midrule
\multirow{5}{*}{VGG16}
 & SVHN          & 96.26 & 95.90 & 2.15 & 3.23 & 33.48\% & 1.35 & 1.67 & 19.23\% \\
 & CIFAR10       & 94.48 & 94.55 & 2.69 & 3.55 & 24.11\% & 1.41 & 1.61 & 12.33\% \\
 & CIFAR100      & 78.15 & 78.65 & 5.87 & 6.90 & 14.97\% & 3.43 & 3.95 & 13.09\% \\
 & ImageNet100R & 88.01 & 87.79 & 8.76 & 9.40 & 6.74\% & 7.66 & 8.31 & 7.83\% \\
 & ImageNet100D & 81.11 & 82.49 & 7.86 & 9.35 & 15.93\% & 7.41 & 8.90 & 16.68\% \\
\midrule
\multirow{5}{*}{ResNet18}
 & SVHN          & 95.43 & 94.36 & 9.06 & 11.07 & 18.23\% & 6.39 & 8.64 & 26.01\% \\
 & CIFAR10       & 94.11 & 94.50 & 8.10 & 11.88 & 31.84\% & 5.52 & 6.95 & 20.61\% \\
 & CIFAR100      & 78.68 & 79.84 & 22.72 & 29.31 & 22.47\% & 14.62 & 21.20 & 31.02\% \\
 & ImageNet100R & 89.37 & 89.56 & 98.20 & 98.44 & 0.25\% & 97.38 & 97.72 & 0.35\% \\
 & ImageNet100D & 84.82 & 84.74 & 99.15 & 99.53 & 0.39\% & 99.15 & 99.53 & 0.39\% \\
\midrule
\multirow{5}{*}{MobileNet}
 & SVHN          & 95.58 & 95.97 & 6.71 & 9.75 & 31.15\% & 3.87 & 5.28 & 26.77\% \\
 & CIFAR10       & 91.90 & 92.43 & 8.89 & 11.22 & 20.76\% & 4.42 & 5.58 & 20.89\% \\
 & CIFAR100      & 65.58 & 66.44 & 13.36 & 16.92 & 21.02\% & 7.67 & 10.39 & 26.20\% \\
 & ImageNet100R & 89.00 & 88.89 & 84.31 & 89.43 & 5.72\% & 79.77 & 85.02 & 6.17\% \\
 & ImageNet100D & 85.10 & 84.43 & 72.73 & 91.57 & 20.57\% & 70.92 & 88.50 & 19.86\% \\
\bottomrule
\end{tabular}}
\vspace{-2mm}
\end{table*}

%% file: sections/7.threats.tex
\section{Threats to Validity}\label{sec:ttv}

\looseness-1
External validity concerns \approach's generalizability to other types of DNN models. %
To mitigate this threat, we used three representative, well-known CNN models of varying architectures and sizes~\cite{vggmodel, resnetmodel, mobilenetmodel}. %
Modularization at the fine-grained neuron level implies the adaptability of \approach to extend to other types of models.

\looseness-1
Internal validity may be affected by a weak research protocol and subject selection bias. We reduced this threat by following well-established practices~\cite{alexnetmodel} and relying on well-known CNN models. Our selected datasets vary in complexity and have been widely used in prior research~\cite{cnnsplitter, mwt, pan2022decomposing}.

\looseness-1
Construct validity rests on the evaluation metrics used. We relied on classification accuracy~\cite{alexnetmodel, vggmodel, resnetmodel, mobilenetmodel} to measure the performance of DNN models. To evaluate the decomposed modules, we evaluate two distinct metrics, i.e., module size (total weights) and module overlap (total shared weights), following a similar approach to previous work~\cite{pan2020decomposing, pan2022decomposing, cnnsplitter}. Additionally, we employed FLOPs~\cite{li2016pruning, luo2020autopruner} to evaluate computational efficiency.

\looseness-1
Finally, conclusion validity concerns the authenticity of the obtained results and fair comparisons between competing methods. 
To mitigate this, we rigorously followed the implementations and default hyper-parameters provided by MwT~\cite{mwt}, GradSplitter~\cite{gradsplitter}, and INCITE~\cite{incite} to reproduce their reported results and apply them to selected models. Note that \approach, MwT, and GradSplitter are implemented in PyTorch, while INCITE uses Keras. We transferred the PyTorch's pre-trained weights to Keras (accuracy difference $\leq$ 0.05\%) and ran INCITE on the converted models. 
For \approach, we rigorously followed the standard ML guidelines~\cite{alexnetmodel} and performed ablation studies to address RQ3.

%% file: sections/8.related_work.tex
\section{Related Work}

\textbf{DNN Modularization} -- Earlier techniques involve analyzing a \textit{fully trained} DNN model to determine which neurons exhibit non-zero activation values when predicting samples of a particular class~\cite{pan2020decomposing, pan2022decomposing, imtiaz2023decomposing, cnnsplitter, qi2023reusing, gradsplitter}. The rationale is that activated neurons are contributing towards the prediction of one class, hence their weights should be grouped into a single module. 
Pan et al.~\cite{pan2020decomposing, pan2022decomposing} use positive–negative sample distinctions to capture those neurons and split a DNN into binary classifiers that can be ensembled through majority voting.
Realizing that not all activated neurons contribute equally to the output, recent work~\cite{cnnsplitter,qi2023reusing,gradsplitter} used a search-based approach to identify minimal subsets of weights that critically impact the predictions of specific classes. 
However, the inherent interconnectivity among hidden units in pre-trained neural networks %
still leads to significant weight overlap between modules~\cite{cnnsplitter, gradsplitter}, hampering their subsequent reuse.
To mitigate the challenge of weight overlap, Qi et al.~\cite{mwt} proposed MwT, which %
enforces a modular structure of a DNN during its training phase by incorporating auxiliary mask generators. 
\approach, on the other hand, aims to promote inherent modularity within DNNs without the need for any such external structure.

Modular DNNs can also be explicitly architected \textit{before} training~\cite{lepikhin2020gshard, andreas2016neural}. For instance, mixture of experts inserts blocks, each containing a sparse gating layer and a predefined set of expert sub-networks, to scale model capacity without increasing computation. This approach demands manual changes to the model architecture. \approach's during-training approach implicitly forms these ``experts'' by promoting modularity within a  DNN, without any architectural modifications. This enables the dynamic, training-time adaptation of each sub-network's structure based on its functional complexity.

\textbf{DNN Reuse} --
\looseness-1
With the growing complexity of DNNs, state-of-the-art models are memory- and computation-intensive for both training and inference~\cite{wang2019deep}.
Research efforts have focused on reducing training costs by reusing task knowledge with transfer learning~\cite{zhuang2020comprehensive, zhao2014online, zhang2011multi}, one/few-shot learning~\cite{fei2006one, song2023comprehensive, sung2018learning} and continual learning~\cite{wang2024comprehensive}. 
Overall, however, they still require the whole model for a new set of tasks.
\approach targets the reusability of specific DNN modules for new tasks, thus potentially curbing training and inference costs~\cite{pan2020decomposing, pan2022decomposing, imtiaz2023decomposing, cnnsplitter, qi2023reusing, gradsplitter, mwt, ren2023deeparc, andreas2016neural, d2021modular, shazeer2017outrageously}.

\textbf{DNN Compression} --
This line of research aims to reduce model inference costs through techniques such as knowledge distillation~\cite{hinton2015distilling, mirzadeh2020improved}, pruning~\cite{luo2020autopruner, li2016pruning}, and quantization~\cite{nagel2021white, hubara2018quantized}. 
DNN pruning is conceptually closest to our work as it identifies unimportant parameters in DNN models, which can reduce a large portion of the parameter matrices~\cite{li2016pruning}.
While \approach also focuses on extracting only relevant weights, it uniquely aims to separate these  weights in terms of individual functionalities, to enable decomposing them into modules and selectively reusing them for new tasks. Conversely, pruning requires repeated efforts to identify and remove irrelevant network parts to address changing requirements.

\textbf{DNN Debugging} --
\looseness-1
Since DNNs are applied in many safety-critical scenarios, several approaches have been proposed to generate test inputs that reveal unexpected behaviors~\cite{pei2017deepxplore, tian2018deeptest, feng2020deepgini, ma2018deepgauge}.
Pei et al.~\cite{pei2017deepxplore} introduced the concept of neuron coverage to pinpoint the parts of a DNNs exercised by a set of test inputs. 
Subsequent work~\cite{tian2018deeptest, feng2020deepgini, ma2018deepgauge} synthesized
test cases with different perturbations to maximize neuron coverage, which can induce erroneous behaviors.
Related research has aimed to formally verify DNNs against different safety properties~\cite{huang2017safety, katz2017reluplex, ehlers2017formal},
and to develop techniques for debugging and repairing DNN models~\cite{eniser2019deepfault, sohn2023arachne, zhang2019apricot, fahmy2021supporting, yu2021deeprepair, duran2021blame}.
Ma et al.~\cite{ma2018mode} analyzed root causes of model misclassifications by identifying faulty neurons based on their output heat maps, and then selecting high quality samples to retrain the models.
Other work~\cite{li2023adaptive,sohn2023arachne} employed search-based methods to adjust neuron weights to improve model accuracy.
\approach is complementary to this line of work since %
the modularity within DNNs enhances their functional segregation, offering the promise of improved interpretability. %

%% file: sections/6.discussion.tex
\section{Discussion  and Conclusions}\label{sec:discussion}

\looseness-1
\approach's  activation-driven, during-training approach, combined with its explicit aim to make the modules compact, has the potential to open new opportunities in the study of DNN modularization and reuse.
Since \approach focuses directly on the activation level of each DNN layer, it is able to achieve greater accuracy while maintaining relative simplicity compared to the state-of-the-art. \approach also keeps training times comparable to the standard baseline, and importantly, does not require retraining or fine-tuning at any point after the fact. %
It is worth noting that \approach carries a one-time training cost before deployment, in return for accuracy-preserving, low‑overlap, compact modules that can be recombined on-the-fly. Post-hoc decomposition techniques may avoid this upfront cost, but their resultant modules inherit the monolithic DNN's entanglement debt. %

\looseness-1
We are actively exploring a range of improvements to \approach. A major goal underlying \approach has been to make it broadly applicable. We have shown that \approach can be applied in essentially the same manner to both convolutional and FC layers. Furthermore, we have applied \approach on different model architectures: VGG's stacked-layer architecture, ResNet's residual block, and MobileNet's depth-wise separable convolutions. We also considered dataset complexity with respect to input dimension (e.g., SVHN/CIFAR-to-ImageNet), sample size (e.g., CIFAR10-to-SVHN), and number of classes (e.g., CIFAR10-to-CIFAR100/ImageNet). 
As with existing work~\cite{cnnsplitter, qi2023reusing, pan2022decomposing,mwt}, to date we have evaluated \approach primarily on CNN models.
Other model families such as RNNs and Transformers are designed to process sequential data across multiple timesteps
, which may involve tailoring of \approach's implementation. We leave this for future work. %

\looseness-1
Another facet of \approach's generalizability is with respect to different activation functions. In \approach, a neuron is active if it has non-zero activation value. Modularity is refined during training by gradually reducing overlapping or unnecessary unit activations toward zero. This broadly applies to various activation functions. As with existing approaches~\cite{pan2020decomposing, pan2022decomposing, qi2023reusing, cnnsplitter, gradsplitter, mwt}, to date we have focused on the commonly-used ReLU. Our future work will extend to other activation functions, such as Leaky ReLU, GELU, Sigmoid, and Tanh. %

\looseness-1
{\approach} enables the reuse and replacement of different parts  of a DNN model without the need to retrain the entire model to meet new requirements. 
To date, we have primarily focused on the reuse of modules for subsets of classes and on replacing existing classes. In general, accuracy-preserving modules produced by {\approach} can also be reused to add new functionalities to a different model. This is an exciting potential application of \approach, but one that may require further adaptation strategies. We will explore this in our future work.

%% file: main.bbl

\begin{thebibliography}{67}


\ifx \showCODEN    \undefined \def \showCODEN     #1{\unskip}     \fi
\ifx \showDOI      \undefined \def \showDOI       #1{#1}\fi
\ifx \showISBNx    \undefined \def \showISBNx     #1{\unskip}     \fi
\ifx \showISBNxiii \undefined \def \showISBNxiii  #1{\unskip}     \fi
\ifx \showISSN     \undefined \def \showISSN      #1{\unskip}     \fi
\ifx \showLCCN     \undefined \def \showLCCN      #1{\unskip}     \fi
\ifx \shownote     \undefined \def \shownote      #1{#1}          \fi
\ifx \showarticletitle \undefined \def \showarticletitle #1{#1}   \fi
\ifx \showURL      \undefined \def \showURL       {\relax}        \fi
\providecommand\bibfield[2]{#2}
\providecommand\bibinfo[2]{#2}
\providecommand\natexlab[1]{#1}
\providecommand\showeprint[2][]{arXiv:#2}

\bibitem[Andreas et~al\mbox{.}(2016)]%
        {andreas2016neural}
\bibfield{author}{\bibinfo{person}{Jacob Andreas}, \bibinfo{person}{Marcus Rohrbach}, \bibinfo{person}{Trevor Darrell}, {and} \bibinfo{person}{Dan Klein}.} \bibinfo{year}{2016}\natexlab{}.
\newblock \showarticletitle{Neural module networks}. In \bibinfo{booktitle}{\emph{Proceedings of the IEEE conference on computer vision and pattern recognition}}. \bibinfo{pages}{39--48}.
\newblock


\bibitem[Ayinde et~al\mbox{.}(2019)]%
        {ayinde2019redundant}
\bibfield{author}{\bibinfo{person}{Babajide~O Ayinde}, \bibinfo{person}{Tamer Inanc}, {and} \bibinfo{person}{Jacek~M Zurada}.} \bibinfo{year}{2019}\natexlab{}.
\newblock \showarticletitle{Redundant feature pruning for accelerated inference in deep neural networks}.
\newblock \bibinfo{journal}{\emph{Neural Networks}}  \bibinfo{volume}{118} (\bibinfo{year}{2019}), \bibinfo{pages}{148--158}.
\newblock


\bibitem[Basha et~al\mbox{.}(2020)]%
        {basha2020impact}
\bibfield{author}{\bibinfo{person}{SH~Shabbeer Basha}, \bibinfo{person}{Shiv~Ram Dubey}, \bibinfo{person}{Viswanath Pulabaigari}, {and} \bibinfo{person}{Snehasis Mukherjee}.} \bibinfo{year}{2020}\natexlab{}.
\newblock \showarticletitle{Impact of fully connected layers on performance of convolutional neural networks for image classification}.
\newblock \bibinfo{journal}{\emph{Neurocomputing}}  \bibinfo{volume}{378} (\bibinfo{year}{2020}), \bibinfo{pages}{112--119}.
\newblock


\bibitem[Bingham and Miikkulainen(2022)]%
        {bingham2022discovering}
\bibfield{author}{\bibinfo{person}{Garrett Bingham} {and} \bibinfo{person}{Risto Miikkulainen}.} \bibinfo{year}{2022}\natexlab{}.
\newblock \showarticletitle{Discovering parametric activation functions}.
\newblock \bibinfo{journal}{\emph{Neural Networks}}  \bibinfo{volume}{148} (\bibinfo{year}{2022}), \bibinfo{pages}{48--65}.
\newblock


\bibitem[Bottou(1998)]%
        {eon1998online}
\bibfield{author}{\bibinfo{person}{L{\'e}on Bottou}.} \bibinfo{year}{1998}\natexlab{}.
\newblock \showarticletitle{Online learning and stochastic approximations}.
\newblock \bibinfo{journal}{\emph{Online learning in neural networks}} \bibinfo{volume}{17}, \bibinfo{number}{9} (\bibinfo{year}{1998}), \bibinfo{pages}{142}.
\newblock


\bibitem[Collobert et~al\mbox{.}(2011)]%
        {collobert2011natural}
\bibfield{author}{\bibinfo{person}{Ronan Collobert}, \bibinfo{person}{Jason Weston}, \bibinfo{person}{L{\'e}on Bottou}, \bibinfo{person}{Michael Karlen}, \bibinfo{person}{Koray Kavukcuoglu}, {and} \bibinfo{person}{Pavel Kuksa}.} \bibinfo{year}{2011}\natexlab{}.
\newblock \showarticletitle{Natural language processing (almost) from scratch}.
\newblock \bibinfo{journal}{\emph{Journal of machine learning research}} \bibinfo{volume}{12}, \bibinfo{number}{ARTICLE} (\bibinfo{year}{2011}), \bibinfo{pages}{2493--2537}.
\newblock


\bibitem[Csord{\'a}s et~al\mbox{.}(2021)]%
        {csordasneural}
\bibfield{author}{\bibinfo{person}{R{\'o}bert Csord{\'a}s}, \bibinfo{person}{Sjoerd van Steenkiste}, {and} \bibinfo{person}{J{\"u}rgen Schmidhuber}.} \bibinfo{year}{2021}\natexlab{}.
\newblock \showarticletitle{Are Neural Nets Modular? Inspecting Functional Modularity Through Differentiable Weight Masks}. In \bibinfo{booktitle}{\emph{International Conference on Learning Representations}}.
\newblock


\bibitem[D'Amario et~al\mbox{.}(2021)]%
        {d2021modular}
\bibfield{author}{\bibinfo{person}{Vanessa D'Amario}, \bibinfo{person}{Tomotake Sasaki}, {and} \bibinfo{person}{Xavier Boix}.} \bibinfo{year}{2021}\natexlab{}.
\newblock \showarticletitle{How modular should neural module networks be for systematic generalization?}
\newblock \bibinfo{journal}{\emph{Advances in Neural Information Processing Systems}}  \bibinfo{volume}{34} (\bibinfo{year}{2021}), \bibinfo{pages}{23374--23385}.
\newblock


\bibitem[Deng et~al\mbox{.}(2009)]%
        {deng2009imagenet}
\bibfield{author}{\bibinfo{person}{Jia Deng}, \bibinfo{person}{Wei Dong}, \bibinfo{person}{Richard Socher}, \bibinfo{person}{Li-Jia Li}, \bibinfo{person}{Kai Li}, {and} \bibinfo{person}{Li Fei-Fei}.} \bibinfo{year}{2009}\natexlab{}.
\newblock \showarticletitle{Imagenet: A large-scale hierarchical image database}. In \bibinfo{booktitle}{\emph{2009 IEEE conference on computer vision and pattern recognition}}. Ieee, \bibinfo{pages}{248--255}.
\newblock


\bibitem[Duran et~al\mbox{.}(2021)]%
        {duran2021blame}
\bibfield{author}{\bibinfo{person}{Matias Duran}, \bibinfo{person}{Xiao-Yi Zhang}, \bibinfo{person}{Paolo Arcaini}, {and} \bibinfo{person}{Fuyuki Ishikawa}.} \bibinfo{year}{2021}\natexlab{}.
\newblock \showarticletitle{What to blame? on the granularity of fault localization for deep neural networks}. In \bibinfo{booktitle}{\emph{2021 IEEE 32nd International Symposium on Software Reliability Engineering (ISSRE)}}. IEEE, \bibinfo{pages}{264--275}.
\newblock


\bibitem[Ehlers(2017)]%
        {ehlers2017formal}
\bibfield{author}{\bibinfo{person}{Ruediger Ehlers}.} \bibinfo{year}{2017}\natexlab{}.
\newblock \showarticletitle{Formal verification of piece-wise linear feed-forward neural networks}. In \bibinfo{booktitle}{\emph{Automated Technology for Verification and Analysis: 15th International Symposium, ATVA 2017, Pune, India, October 3--6, 2017, Proceedings 15}}. Springer, \bibinfo{pages}{269--286}.
\newblock


\bibitem[Eniser et~al\mbox{.}(2019)]%
        {eniser2019deepfault}
\bibfield{author}{\bibinfo{person}{Hasan~Ferit Eniser}, \bibinfo{person}{Simos Gerasimou}, {and} \bibinfo{person}{Alper Sen}.} \bibinfo{year}{2019}\natexlab{}.
\newblock \showarticletitle{Deepfault: Fault localization for deep neural networks}. In \bibinfo{booktitle}{\emph{International Conference on Fundamental Approaches to Software Engineering}}. Springer, \bibinfo{pages}{171--191}.
\newblock


\bibitem[Fahmy et~al\mbox{.}(2021)]%
        {fahmy2021supporting}
\bibfield{author}{\bibinfo{person}{Hazem Fahmy}, \bibinfo{person}{Fabrizio Pastore}, \bibinfo{person}{Mojtaba Bagherzadeh}, {and} \bibinfo{person}{Lionel Briand}.} \bibinfo{year}{2021}\natexlab{}.
\newblock \showarticletitle{Supporting deep neural network safety analysis and retraining through heatmap-based unsupervised learning}.
\newblock \bibinfo{journal}{\emph{IEEE Transactions on Reliability}} \bibinfo{volume}{70}, \bibinfo{number}{4} (\bibinfo{year}{2021}), \bibinfo{pages}{1641--1657}.
\newblock


\bibitem[Fei-Fei et~al\mbox{.}(2006)]%
        {fei2006one}
\bibfield{author}{\bibinfo{person}{Li Fei-Fei}, \bibinfo{person}{Robert Fergus}, {and} \bibinfo{person}{Pietro Perona}.} \bibinfo{year}{2006}\natexlab{}.
\newblock \showarticletitle{One-shot learning of object categories}.
\newblock \bibinfo{journal}{\emph{IEEE transactions on pattern analysis and machine intelligence}} \bibinfo{volume}{28}, \bibinfo{number}{4} (\bibinfo{year}{2006}), \bibinfo{pages}{594--611}.
\newblock


\bibitem[Feng et~al\mbox{.}(2020)]%
        {feng2020deepgini}
\bibfield{author}{\bibinfo{person}{Yang Feng}, \bibinfo{person}{Qingkai Shi}, \bibinfo{person}{Xinyu Gao}, \bibinfo{person}{Jun Wan}, \bibinfo{person}{Chunrong Fang}, {and} \bibinfo{person}{Zhenyu Chen}.} \bibinfo{year}{2020}\natexlab{}.
\newblock \showarticletitle{Deepgini: prioritizing massive tests to enhance the robustness of deep neural networks}. In \bibinfo{booktitle}{\emph{Proceedings of the 29th ACM SIGSOFT International Symposium on Software Testing and Analysis}}. \bibinfo{pages}{177--188}.
\newblock


\bibitem[Ghanbari(2024)]%
        {incite}
\bibfield{author}{\bibinfo{person}{Ali Ghanbari}.} \bibinfo{year}{2024}\natexlab{}.
\newblock \showarticletitle{Decomposition of deep neural networks into modules via mutation analysis}. In \bibinfo{booktitle}{\emph{Proceedings of the 33rd ACM SIGSOFT International Symposium on Software Testing and Analysis}}. \bibinfo{pages}{1669--1681}.
\newblock


\bibitem[He et~al\mbox{.}(2016)]%
        {resnetmodel}
\bibfield{author}{\bibinfo{person}{Kaiming He}, \bibinfo{person}{Xiangyu Zhang}, \bibinfo{person}{Shaoqing Ren}, {and} \bibinfo{person}{Jian Sun}.} \bibinfo{year}{2016}\natexlab{}.
\newblock \showarticletitle{Deep residual learning for image recognition}. In \bibinfo{booktitle}{\emph{Proceedings of the IEEE conference on computer vision and pattern recognition}}. \bibinfo{pages}{770--778}.
\newblock


\bibitem[Hinton et~al\mbox{.}(2015)]%
        {hinton2015distilling}
\bibfield{author}{\bibinfo{person}{Geoffrey Hinton}, \bibinfo{person}{Oriol Vinyals}, {and} \bibinfo{person}{Jeff Dean}.} \bibinfo{year}{2015}\natexlab{}.
\newblock \showarticletitle{Distilling the knowledge in a neural network}.
\newblock \bibinfo{journal}{\emph{NIPS Deep Learning and Representation Learning Workshop}} (\bibinfo{year}{2015}).
\newblock


\bibitem[Howard et~al\mbox{.}(2017)]%
        {mobilenetmodel}
\bibfield{author}{\bibinfo{person}{Andrew~G Howard}, \bibinfo{person}{Menglong Zhu}, \bibinfo{person}{Bo Chen}, \bibinfo{person}{Dmitry Kalenichenko}, \bibinfo{person}{Weijun Wang}, \bibinfo{person}{Tobias Weyand}, \bibinfo{person}{Marco Andreetto}, {and} \bibinfo{person}{Hartwig Adam}.} \bibinfo{year}{2017}\natexlab{}.
\newblock \showarticletitle{Mobilenets: Efficient convolutional neural networks for mobile vision applications}.
\newblock \bibinfo{journal}{\emph{arXiv preprint arXiv:1704.04861}} (\bibinfo{year}{2017}).
\newblock


\bibitem[Huang et~al\mbox{.}(2017)]%
        {huang2017safety}
\bibfield{author}{\bibinfo{person}{Xiaowei Huang}, \bibinfo{person}{Marta Kwiatkowska}, \bibinfo{person}{Sen Wang}, {and} \bibinfo{person}{Min Wu}.} \bibinfo{year}{2017}\natexlab{}.
\newblock \showarticletitle{Safety verification of deep neural networks}. In \bibinfo{booktitle}{\emph{Computer Aided Verification: 29th International Conference, CAV 2017, Heidelberg, Germany, July 24-28, 2017, Proceedings, Part I 30}}. Springer, \bibinfo{pages}{3--29}.
\newblock


\bibitem[Hubara et~al\mbox{.}(2018)]%
        {hubara2018quantized}
\bibfield{author}{\bibinfo{person}{Itay Hubara}, \bibinfo{person}{Matthieu Courbariaux}, \bibinfo{person}{Daniel Soudry}, \bibinfo{person}{Ran El-Yaniv}, {and} \bibinfo{person}{Yoshua Bengio}.} \bibinfo{year}{2018}\natexlab{}.
\newblock \showarticletitle{Quantized neural networks: Training neural networks with low precision weights and activations}.
\newblock \bibinfo{journal}{\emph{Journal of Machine Learning Research}} \bibinfo{volume}{18}, \bibinfo{number}{187} (\bibinfo{year}{2018}), \bibinfo{pages}{1--30}.
\newblock


\bibitem[Imtiaz et~al\mbox{.}(2023)]%
        {imtiaz2023decomposing}
\bibfield{author}{\bibinfo{person}{Sayem~Mohammad Imtiaz}, \bibinfo{person}{Fraol Batole}, \bibinfo{person}{Astha Singh}, \bibinfo{person}{Rangeet Pan}, \bibinfo{person}{Breno~Dantas Cruz}, {and} \bibinfo{person}{Hridesh Rajan}.} \bibinfo{year}{2023}\natexlab{}.
\newblock \showarticletitle{Decomposing a recurrent neural network into modules for enabling reusability and replacement}. In \bibinfo{booktitle}{\emph{2023 IEEE/ACM 45th International Conference on Software Engineering (ICSE)}}. IEEE, \bibinfo{pages}{1020--1032}.
\newblock


\bibitem[Katz et~al\mbox{.}(2017)]%
        {katz2017reluplex}
\bibfield{author}{\bibinfo{person}{Guy Katz}, \bibinfo{person}{Clark Barrett}, \bibinfo{person}{David~L Dill}, \bibinfo{person}{Kyle Julian}, {and} \bibinfo{person}{Mykel~J Kochenderfer}.} \bibinfo{year}{2017}\natexlab{}.
\newblock \showarticletitle{Reluplex: An efficient SMT solver for verifying deep neural networks}. In \bibinfo{booktitle}{\emph{Computer Aided Verification: 29th International Conference, CAV 2017, Heidelberg, Germany, July 24-28, 2017, Proceedings, Part I 30}}. Springer, \bibinfo{pages}{97--117}.
\newblock


\bibitem[Krizhevsky et~al\mbox{.}(2009)]%
        {cifar10dataset}
\bibfield{author}{\bibinfo{person}{Alex Krizhevsky}, \bibinfo{person}{Geoffrey Hinton}, {et~al\mbox{.}}} \bibinfo{year}{2009}\natexlab{}.
\newblock \showarticletitle{Learning multiple layers of features from tiny images}.
\newblock  (\bibinfo{year}{2009}).
\newblock
\newblock
\shownote{Technical report}.


\bibitem[Krizhevsky et~al\mbox{.}(2012)]%
        {alexnetmodel}
\bibfield{author}{\bibinfo{person}{Alex Krizhevsky}, \bibinfo{person}{Ilya Sutskever}, {and} \bibinfo{person}{Geoffrey~E Hinton}.} \bibinfo{year}{2012}\natexlab{}.
\newblock \showarticletitle{Imagenet classification with deep convolutional neural networks}.
\newblock \bibinfo{journal}{\emph{Advances in neural information processing systems}}  \bibinfo{volume}{25} (\bibinfo{year}{2012}).
\newblock


\bibitem[Krogh and Hertz(1991)]%
        {krogh1991simple}
\bibfield{author}{\bibinfo{person}{Anders Krogh} {and} \bibinfo{person}{John Hertz}.} \bibinfo{year}{1991}\natexlab{}.
\newblock \showarticletitle{A simple weight decay can improve generalization}.
\newblock \bibinfo{journal}{\emph{Advances in neural information processing systems}}  \bibinfo{volume}{4} (\bibinfo{year}{1991}).
\newblock


\bibitem[LeCun et~al\mbox{.}(1998)]%
        {lenet5}
\bibfield{author}{\bibinfo{person}{Yann LeCun}, \bibinfo{person}{L{\'e}on Bottou}, \bibinfo{person}{Yoshua Bengio}, {and} \bibinfo{person}{Patrick Haffner}.} \bibinfo{year}{1998}\natexlab{}.
\newblock \showarticletitle{Gradient-based learning applied to document recognition}.
\newblock \bibinfo{journal}{\emph{Proc. IEEE}} \bibinfo{volume}{86}, \bibinfo{number}{11} (\bibinfo{year}{1998}), \bibinfo{pages}{2278--2324}.
\newblock


\bibitem[Lepikhin et~al\mbox{.}(2021)]%
        {lepikhin2020gshard}
\bibfield{author}{\bibinfo{person}{Dmitry Lepikhin}, \bibinfo{person}{HyoukJoong Lee}, \bibinfo{person}{Yuanzhong Xu}, \bibinfo{person}{Dehao Chen}, \bibinfo{person}{Orhan Firat}, \bibinfo{person}{Yanping Huang}, \bibinfo{person}{Maxim Krikun}, \bibinfo{person}{Noam Shazeer}, {and} \bibinfo{person}{Zhifeng Chen}.} \bibinfo{year}{2021}\natexlab{}.
\newblock \showarticletitle{Gshard: Scaling giant models with conditional computation and automatic sharding}.
\newblock \bibinfo{journal}{\emph{International Conference on Learning Representations}} (\bibinfo{year}{2021}).
\newblock


\bibitem[Li et~al\mbox{.}(2017)]%
        {li2016pruning}
\bibfield{author}{\bibinfo{person}{Hao Li}, \bibinfo{person}{Asim Kadav}, \bibinfo{person}{Igor Durdanovic}, \bibinfo{person}{Hanan Samet}, {and} \bibinfo{person}{Hans~Peter Graf}.} \bibinfo{year}{2017}\natexlab{}.
\newblock \showarticletitle{Pruning filters for efficient convnets}.
\newblock \bibinfo{journal}{\emph{International Conference on Learning Representations}} (\bibinfo{year}{2017}).
\newblock


\bibitem[Li~Calsi et~al\mbox{.}(2023)]%
        {li2023adaptive}
\bibfield{author}{\bibinfo{person}{Davide Li~Calsi}, \bibinfo{person}{Matias Duran}, \bibinfo{person}{Thomas Laurent}, \bibinfo{person}{Xiao-Yi Zhang}, \bibinfo{person}{Paolo Arcaini}, {and} \bibinfo{person}{Fuyuki Ishikawa}.} \bibinfo{year}{2023}\natexlab{}.
\newblock \showarticletitle{Adaptive search-based repair of deep neural networks}. In \bibinfo{booktitle}{\emph{Proceedings of the Genetic and Evolutionary Computation Conference}}. \bibinfo{pages}{1527--1536}.
\newblock


\bibitem[Liu et~al\mbox{.}(2016)]%
        {liu2016towards}
\bibfield{author}{\bibinfo{person}{Mengchen Liu}, \bibinfo{person}{Jiaxin Shi}, \bibinfo{person}{Zhen Li}, \bibinfo{person}{Chongxuan Li}, \bibinfo{person}{Jun Zhu}, {and} \bibinfo{person}{Shixia Liu}.} \bibinfo{year}{2016}\natexlab{}.
\newblock \showarticletitle{Towards better analysis of deep convolutional neural networks}.
\newblock \bibinfo{journal}{\emph{IEEE transactions on visualization and computer graphics}} \bibinfo{volume}{23}, \bibinfo{number}{1} (\bibinfo{year}{2016}), \bibinfo{pages}{91--100}.
\newblock


\bibitem[Luo and Wu(2020)]%
        {luo2020autopruner}
\bibfield{author}{\bibinfo{person}{Jian-Hao Luo} {and} \bibinfo{person}{Jianxin Wu}.} \bibinfo{year}{2020}\natexlab{}.
\newblock \showarticletitle{Autopruner: An end-to-end trainable filter pruning method for efficient deep model inference}.
\newblock \bibinfo{journal}{\emph{Pattern Recognition}}  \bibinfo{volume}{107} (\bibinfo{year}{2020}), \bibinfo{pages}{107461}.
\newblock


\bibitem[Ma et~al\mbox{.}(2018a)]%
        {ma2018deepgauge}
\bibfield{author}{\bibinfo{person}{Lei Ma}, \bibinfo{person}{Felix Juefei-Xu}, \bibinfo{person}{Fuyuan Zhang}, \bibinfo{person}{Jiyuan Sun}, \bibinfo{person}{Minhui Xue}, \bibinfo{person}{Bo Li}, \bibinfo{person}{Chunyang Chen}, \bibinfo{person}{Ting Su}, \bibinfo{person}{Li Li}, \bibinfo{person}{Yang Liu}, {et~al\mbox{.}}} \bibinfo{year}{2018}\natexlab{a}.
\newblock \showarticletitle{Deepgauge: Multi-granularity testing criteria for deep learning systems}. In \bibinfo{booktitle}{\emph{Proceedings of the 33rd ACM/IEEE international conference on automated software engineering}}. \bibinfo{pages}{120--131}.
\newblock


\bibitem[Ma et~al\mbox{.}(2019)]%
        {ma2019transformed}
\bibfield{author}{\bibinfo{person}{Rongrong Ma}, \bibinfo{person}{Jianyu Miao}, \bibinfo{person}{Lingfeng Niu}, {and} \bibinfo{person}{Peng Zhang}.} \bibinfo{year}{2019}\natexlab{}.
\newblock \showarticletitle{Transformed L1 regularization for learning sparse deep neural networks}.
\newblock \bibinfo{journal}{\emph{Neural Networks}}  \bibinfo{volume}{119} (\bibinfo{year}{2019}), \bibinfo{pages}{286--298}.
\newblock


\bibitem[Ma et~al\mbox{.}(2018b)]%
        {ma2018mode}
\bibfield{author}{\bibinfo{person}{Shiqing Ma}, \bibinfo{person}{Yingqi Liu}, \bibinfo{person}{Wen-Chuan Lee}, \bibinfo{person}{Xiangyu Zhang}, {and} \bibinfo{person}{Ananth Grama}.} \bibinfo{year}{2018}\natexlab{b}.
\newblock \showarticletitle{MODE: automated neural network model debugging via state differential analysis and input selection}. In \bibinfo{booktitle}{\emph{Proceedings of the 2018 26th ACM Joint Meeting on European Software Engineering Conference and Symposium on the Foundations of Software Engineering}}. \bibinfo{pages}{175--186}.
\newblock


\bibitem[McCloskey and Cohen(1989)]%
        {mccloskey1989catastrophic}
\bibfield{author}{\bibinfo{person}{Michael McCloskey} {and} \bibinfo{person}{Neal~J Cohen}.} \bibinfo{year}{1989}\natexlab{}.
\newblock \showarticletitle{Catastrophic interference in connectionist networks: The sequential learning problem}.
\newblock In \bibinfo{booktitle}{\emph{Psychology of learning and motivation}}. Vol.~\bibinfo{volume}{24}. \bibinfo{publisher}{Elsevier}, \bibinfo{pages}{109--165}.
\newblock


\bibitem[Mirzadeh et~al\mbox{.}(2020)]%
        {mirzadeh2020improved}
\bibfield{author}{\bibinfo{person}{Seyed~Iman Mirzadeh}, \bibinfo{person}{Mehrdad Farajtabar}, \bibinfo{person}{Ang Li}, \bibinfo{person}{Nir Levine}, \bibinfo{person}{Akihiro Matsukawa}, {and} \bibinfo{person}{Hassan Ghasemzadeh}.} \bibinfo{year}{2020}\natexlab{}.
\newblock \showarticletitle{Improved knowledge distillation via teacher assistant}. In \bibinfo{booktitle}{\emph{Proceedings of the AAAI conference on artificial intelligence}}, Vol.~\bibinfo{volume}{34}. \bibinfo{pages}{5191--5198}.
\newblock


\bibitem[Nagel et~al\mbox{.}(2021)]%
        {nagel2021white}
\bibfield{author}{\bibinfo{person}{Markus Nagel}, \bibinfo{person}{Marios Fournarakis}, \bibinfo{person}{Rana~Ali Amjad}, \bibinfo{person}{Yelysei Bondarenko}, \bibinfo{person}{Mart Van~Baalen}, {and} \bibinfo{person}{Tijmen Blankevoort}.} \bibinfo{year}{2021}\natexlab{}.
\newblock \showarticletitle{A white paper on neural network quantization}.
\newblock \bibinfo{journal}{\emph{arXiv preprint arXiv:2106.08295}} (\bibinfo{year}{2021}).
\newblock


\bibitem[Nassif et~al\mbox{.}(2019)]%
        {nassif2019speech}
\bibfield{author}{\bibinfo{person}{Ali~Bou Nassif}, \bibinfo{person}{Ismail Shahin}, \bibinfo{person}{Imtinan Attili}, \bibinfo{person}{Mohammad Azzeh}, {and} \bibinfo{person}{Khaled Shaalan}.} \bibinfo{year}{2019}\natexlab{}.
\newblock \showarticletitle{Speech recognition using deep neural networks: A systematic review}.
\newblock \bibinfo{journal}{\emph{IEEE access}}  \bibinfo{volume}{7} (\bibinfo{year}{2019}), \bibinfo{pages}{19143--19165}.
\newblock


\bibitem[Netzer et~al\mbox{.}(2011)]%
        {svhndataset}
\bibfield{author}{\bibinfo{person}{Yuval Netzer}, \bibinfo{person}{Tao Wang}, \bibinfo{person}{Adam Coates}, \bibinfo{person}{Alessandro Bissacco}, \bibinfo{person}{Baolin Wu}, \bibinfo{person}{Andrew~Y Ng}, {et~al\mbox{.}}} \bibinfo{year}{2011}\natexlab{}.
\newblock \showarticletitle{Reading digits in natural images with unsupervised feature learning}. In \bibinfo{booktitle}{\emph{NIPS workshop on deep learning and unsupervised feature learning}}, Vol.~\bibinfo{volume}{2011}. Granada, \bibinfo{pages}{7}.
\newblock


\bibitem[Ngo et~al\mbox{.}(2025)]%
        {website}
\bibfield{author}{\bibinfo{person}{Tuan Ngo}, \bibinfo{person}{Abid Hassan}, \bibinfo{person}{Saad Shafiq}, {and} \bibinfo{person}{Nenad Medvidovi\'c}.} \bibinfo{year}{2025}\natexlab{}.
\newblock \bibinfo{title}{{MODA}}.
\newblock
\newblock
\urldef\tempurl%
\url{https://sites.google.com/view/dnn-moda}
\showURL{%
\tempurl}


\bibitem[Pan and Rajan(2020)]%
        {pan2020decomposing}
\bibfield{author}{\bibinfo{person}{Rangeet Pan} {and} \bibinfo{person}{Hridesh Rajan}.} \bibinfo{year}{2020}\natexlab{}.
\newblock \showarticletitle{On decomposing a deep neural network into modules}. In \bibinfo{booktitle}{\emph{Proceedings of the 28th ACM Joint Meeting on European Software Engineering Conference and Symposium on the Foundations of Software Engineering}}. \bibinfo{pages}{889--900}.
\newblock


\bibitem[Pan and Rajan(2022)]%
        {pan2022decomposing}
\bibfield{author}{\bibinfo{person}{Rangeet Pan} {and} \bibinfo{person}{Hridesh Rajan}.} \bibinfo{year}{2022}\natexlab{}.
\newblock \showarticletitle{Decomposing convolutional neural networks into reusable and replaceable modules}. In \bibinfo{booktitle}{\emph{Proceedings of the 44th International Conference on Software Engineering}}. \bibinfo{pages}{524--535}.
\newblock


\bibitem[Parnas(1972)]%
        {parnas1972criteria}
\bibfield{author}{\bibinfo{person}{David~Lorge Parnas}.} \bibinfo{year}{1972}\natexlab{}.
\newblock \showarticletitle{On the criteria to be used in decomposing systems into modules}.
\newblock \bibinfo{journal}{\emph{Commun. ACM}} \bibinfo{volume}{15}, \bibinfo{number}{12} (\bibinfo{year}{1972}), \bibinfo{pages}{1053--1058}.
\newblock


\bibitem[Parnas(1976)]%
        {parnas1976design}
\bibfield{author}{\bibinfo{person}{David~Lorge Parnas}.} \bibinfo{year}{1976}\natexlab{}.
\newblock \showarticletitle{On the design and development of program families}.
\newblock \bibinfo{journal}{\emph{IEEE Transactions on software engineering}} \bibinfo{number}{1} (\bibinfo{year}{1976}), \bibinfo{pages}{1--9}.
\newblock


\bibitem[Pei et~al\mbox{.}(2017)]%
        {pei2017deepxplore}
\bibfield{author}{\bibinfo{person}{Kexin Pei}, \bibinfo{person}{Yinzhi Cao}, \bibinfo{person}{Junfeng Yang}, {and} \bibinfo{person}{Suman Jana}.} \bibinfo{year}{2017}\natexlab{}.
\newblock \showarticletitle{Deepxplore: Automated whitebox testing of deep learning systems}. In \bibinfo{booktitle}{\emph{proceedings of the 26th Symposium on Operating Systems Principles}}. \bibinfo{pages}{1--18}.
\newblock


\bibitem[Qi et~al\mbox{.}(2022)]%
        {cnnsplitter}
\bibfield{author}{\bibinfo{person}{Binhang Qi}, \bibinfo{person}{Hailong Sun}, \bibinfo{person}{Xiang Gao}, {and} \bibinfo{person}{Hongyu Zhang}.} \bibinfo{year}{2022}\natexlab{}.
\newblock \showarticletitle{Patching weak convolutional neural network models through modularization and composition}. In \bibinfo{booktitle}{\emph{Proceedings of the 37th IEEE/ACM International Conference on Automated Software Engineering}}. \bibinfo{pages}{1--12}.
\newblock


\bibitem[Qi et~al\mbox{.}(2023a)]%
        {qi2023reusing}
\bibfield{author}{\bibinfo{person}{Binhang Qi}, \bibinfo{person}{Hailong Sun}, \bibinfo{person}{Xiang Gao}, \bibinfo{person}{Hongyu Zhang}, \bibinfo{person}{Zhaotian Li}, {and} \bibinfo{person}{Xudong Liu}.} \bibinfo{year}{2023}\natexlab{a}.
\newblock \showarticletitle{Reusing deep neural network models through model re-engineering}. In \bibinfo{booktitle}{\emph{2023 IEEE/ACM 45th International Conference on Software Engineering (ICSE)}}. IEEE, \bibinfo{pages}{983--994}.
\newblock


\bibitem[Qi et~al\mbox{.}(2023b)]%
        {gradsplitter}
\bibfield{author}{\bibinfo{person}{Binhang Qi}, \bibinfo{person}{Hailong Sun}, \bibinfo{person}{Hongyu Zhang}, {and} \bibinfo{person}{Xiang Gao}.} \bibinfo{year}{2023}\natexlab{b}.
\newblock \showarticletitle{Reusing Convolutional Neural Network Models through Modularization and Composition}.
\newblock \bibinfo{journal}{\emph{ACM Transactions on Software Engineering and Methodology}} (\bibinfo{year}{2023}).
\newblock


\bibitem[Qi et~al\mbox{.}(2024)]%
        {mwt}
\bibfield{author}{\bibinfo{person}{Binhang Qi}, \bibinfo{person}{Hailong Sun}, \bibinfo{person}{Hongyu Zhang}, \bibinfo{person}{Ruobing Zhao}, {and} \bibinfo{person}{Xiang Gao}.} \bibinfo{year}{2024}\natexlab{}.
\newblock \showarticletitle{Modularizing while Training: A New Paradigm for Modularizing DNN Models}. In \bibinfo{booktitle}{\emph{Proceedings of the 46th IEEE/ACM International Conference on Software Engineering}}. \bibinfo{pages}{1--12}.
\newblock


\bibitem[Ren et~al\mbox{.}(2023)]%
        {ren2023deeparc}
\bibfield{author}{\bibinfo{person}{Xiaoning Ren}, \bibinfo{person}{Yun Lin}, \bibinfo{person}{Yinxing Xue}, \bibinfo{person}{Ruofan Liu}, \bibinfo{person}{Jun Sun}, \bibinfo{person}{Zhiyong Feng}, {and} \bibinfo{person}{Jin~Song Dong}.} \bibinfo{year}{2023}\natexlab{}.
\newblock \showarticletitle{Deeparc: Modularizing neural networks for the model maintenance}. In \bibinfo{booktitle}{\emph{2023 IEEE/ACM 45th International Conference on Software Engineering (ICSE)}}. IEEE, \bibinfo{pages}{1008--1019}.
\newblock


\bibitem[Sculley et~al\mbox{.}(2015)]%
        {sculley2015hidden}
\bibfield{author}{\bibinfo{person}{David Sculley}, \bibinfo{person}{Gary Holt}, \bibinfo{person}{Daniel Golovin}, \bibinfo{person}{Eugene Davydov}, \bibinfo{person}{Todd Phillips}, \bibinfo{person}{Dietmar Ebner}, \bibinfo{person}{Vinay Chaudhary}, \bibinfo{person}{Michael Young}, \bibinfo{person}{Jean-Francois Crespo}, {and} \bibinfo{person}{Dan Dennison}.} \bibinfo{year}{2015}\natexlab{}.
\newblock \showarticletitle{Hidden technical debt in machine learning systems}.
\newblock \bibinfo{journal}{\emph{Advances in neural information processing systems}}  \bibinfo{volume}{28} (\bibinfo{year}{2015}).
\newblock


\bibitem[Shazeer et~al\mbox{.}(2017)]%
        {shazeer2017outrageously}
\bibfield{author}{\bibinfo{person}{Noam Shazeer}, \bibinfo{person}{Azalia Mirhoseini}, \bibinfo{person}{Krzysztof Maziarz}, \bibinfo{person}{Andy Davis}, \bibinfo{person}{Quoc Le}, \bibinfo{person}{Geoffrey Hinton}, {and} \bibinfo{person}{Jeff Dean}.} \bibinfo{year}{2017}\natexlab{}.
\newblock \showarticletitle{Outrageously large neural networks: The sparsely-gated mixture-of-experts layer}.
\newblock \bibinfo{journal}{\emph{International Conference on Learning Representations}} (\bibinfo{year}{2017}).
\newblock


\bibitem[Shorten and Khoshgoftaar(2019)]%
        {shorten2019survey}
\bibfield{author}{\bibinfo{person}{Connor Shorten} {and} \bibinfo{person}{Taghi~M Khoshgoftaar}.} \bibinfo{year}{2019}\natexlab{}.
\newblock \showarticletitle{A survey on image data augmentation for deep learning}.
\newblock \bibinfo{journal}{\emph{Journal of big data}} \bibinfo{volume}{6}, \bibinfo{number}{1} (\bibinfo{year}{2019}), \bibinfo{pages}{1--48}.
\newblock


\bibitem[Simonyan and Zisserman(2015)]%
        {vggmodel}
\bibfield{author}{\bibinfo{person}{K Simonyan} {and} \bibinfo{person}{A Zisserman}.} \bibinfo{year}{2015}\natexlab{}.
\newblock \showarticletitle{Very deep convolutional networks for large-scale image recognition}. In \bibinfo{booktitle}{\emph{3rd International Conference on Learning Representations (ICLR 2015)}}. Computational and Biological Learning Society.
\newblock


\bibitem[Sohn et~al\mbox{.}(2023)]%
        {sohn2023arachne}
\bibfield{author}{\bibinfo{person}{Jeongju Sohn}, \bibinfo{person}{Sungmin Kang}, {and} \bibinfo{person}{Shin Yoo}.} \bibinfo{year}{2023}\natexlab{}.
\newblock \showarticletitle{Arachne: Search-based repair of deep neural networks}.
\newblock \bibinfo{journal}{\emph{ACM Transactions on Software Engineering and Methodology}} \bibinfo{volume}{32}, \bibinfo{number}{4} (\bibinfo{year}{2023}), \bibinfo{pages}{1--26}.
\newblock


\bibitem[Song et~al\mbox{.}(2023)]%
        {song2023comprehensive}
\bibfield{author}{\bibinfo{person}{Yisheng Song}, \bibinfo{person}{Ting Wang}, \bibinfo{person}{Puyu Cai}, \bibinfo{person}{Subrota~K Mondal}, {and} \bibinfo{person}{Jyoti~Prakash Sahoo}.} \bibinfo{year}{2023}\natexlab{}.
\newblock \showarticletitle{A comprehensive survey of few-shot learning: Evolution, applications, challenges, and opportunities}.
\newblock \bibinfo{journal}{\emph{Comput. Surveys}} (\bibinfo{year}{2023}).
\newblock


\bibitem[Srivastava et~al\mbox{.}(2014)]%
        {srivastava2014dropout}
\bibfield{author}{\bibinfo{person}{Nitish Srivastava}, \bibinfo{person}{Geoffrey Hinton}, \bibinfo{person}{Alex Krizhevsky}, \bibinfo{person}{Ilya Sutskever}, {and} \bibinfo{person}{Ruslan Salakhutdinov}.} \bibinfo{year}{2014}\natexlab{}.
\newblock \showarticletitle{Dropout: a simple way to prevent neural networks from overfitting}.
\newblock \bibinfo{journal}{\emph{The journal of machine learning research}} \bibinfo{volume}{15}, \bibinfo{number}{1} (\bibinfo{year}{2014}), \bibinfo{pages}{1929--1958}.
\newblock


\bibitem[Sung et~al\mbox{.}(2018)]%
        {sung2018learning}
\bibfield{author}{\bibinfo{person}{Flood Sung}, \bibinfo{person}{Yongxin Yang}, \bibinfo{person}{Li Zhang}, \bibinfo{person}{Tao Xiang}, \bibinfo{person}{Philip~HS Torr}, {and} \bibinfo{person}{Timothy~M Hospedales}.} \bibinfo{year}{2018}\natexlab{}.
\newblock \showarticletitle{Learning to compare: Relation network for few-shot learning}. In \bibinfo{booktitle}{\emph{Proceedings of the IEEE conference on computer vision and pattern recognition}}. \bibinfo{pages}{1199--1208}.
\newblock


\bibitem[Tian et~al\mbox{.}(2018)]%
        {tian2018deeptest}
\bibfield{author}{\bibinfo{person}{Yuchi Tian}, \bibinfo{person}{Kexin Pei}, \bibinfo{person}{Suman Jana}, {and} \bibinfo{person}{Baishakhi Ray}.} \bibinfo{year}{2018}\natexlab{}.
\newblock \showarticletitle{Deeptest: Automated testing of deep-neural-network-driven autonomous cars}. In \bibinfo{booktitle}{\emph{Proceedings of the 40th international conference on software engineering}}. \bibinfo{pages}{303--314}.
\newblock


\bibitem[Wang et~al\mbox{.}(2019)]%
        {wang2019deep}
\bibfield{author}{\bibinfo{person}{Erwei Wang}, \bibinfo{person}{James~J Davis}, \bibinfo{person}{Ruizhe Zhao}, \bibinfo{person}{Ho-Cheung Ng}, \bibinfo{person}{Xinyu Niu}, \bibinfo{person}{Wayne Luk}, \bibinfo{person}{Peter~YK Cheung}, {and} \bibinfo{person}{George~A Constantinides}.} \bibinfo{year}{2019}\natexlab{}.
\newblock \showarticletitle{Deep neural network approximation for custom hardware: Where we've been, where we're going}.
\newblock \bibinfo{journal}{\emph{ACM Computing Surveys (CSUR)}} \bibinfo{volume}{52}, \bibinfo{number}{2} (\bibinfo{year}{2019}), \bibinfo{pages}{1--39}.
\newblock


\bibitem[Wang et~al\mbox{.}(2024)]%
        {wang2024comprehensive}
\bibfield{author}{\bibinfo{person}{Liyuan Wang}, \bibinfo{person}{Xingxing Zhang}, \bibinfo{person}{Hang Su}, {and} \bibinfo{person}{Jun Zhu}.} \bibinfo{year}{2024}\natexlab{}.
\newblock \showarticletitle{A comprehensive survey of continual learning: Theory, method and application}.
\newblock \bibinfo{journal}{\emph{IEEE Transactions on Pattern Analysis and Machine Intelligence}} (\bibinfo{year}{2024}).
\newblock


\bibitem[Yu et~al\mbox{.}(2021)]%
        {yu2021deeprepair}
\bibfield{author}{\bibinfo{person}{Bing Yu}, \bibinfo{person}{Hua Qi}, \bibinfo{person}{Qing Guo}, \bibinfo{person}{Felix Juefei-Xu}, \bibinfo{person}{Xiaofei Xie}, \bibinfo{person}{Lei Ma}, {and} \bibinfo{person}{Jianjun Zhao}.} \bibinfo{year}{2021}\natexlab{}.
\newblock \showarticletitle{Deeprepair: Style-guided repairing for deep neural networks in the real-world operational environment}.
\newblock \bibinfo{journal}{\emph{IEEE Transactions on Reliability}} \bibinfo{volume}{71}, \bibinfo{number}{4} (\bibinfo{year}{2021}), \bibinfo{pages}{1401--1416}.
\newblock


\bibitem[Zhang et~al\mbox{.}(2011)]%
        {zhang2011multi}
\bibfield{author}{\bibinfo{person}{Dan Zhang}, \bibinfo{person}{Jingrui He}, \bibinfo{person}{Yan Liu}, \bibinfo{person}{Luo Si}, {and} \bibinfo{person}{Richard Lawrence}.} \bibinfo{year}{2011}\natexlab{}.
\newblock \showarticletitle{Multi-view transfer learning with a large margin approach}. In \bibinfo{booktitle}{\emph{Proceedings of the 17th ACM SIGKDD international conference on Knowledge discovery and data mining}}. \bibinfo{pages}{1208--1216}.
\newblock


\bibitem[Zhang and Chan(2019)]%
        {zhang2019apricot}
\bibfield{author}{\bibinfo{person}{Hao Zhang} {and} \bibinfo{person}{WK Chan}.} \bibinfo{year}{2019}\natexlab{}.
\newblock \showarticletitle{Apricot: A weight-adaptation approach to fixing deep learning models}. In \bibinfo{booktitle}{\emph{2019 34th IEEE/ACM International Conference on Automated Software Engineering (ASE)}}. IEEE, \bibinfo{pages}{376--387}.
\newblock


\bibitem[Zhao et~al\mbox{.}(2014)]%
        {zhao2014online}
\bibfield{author}{\bibinfo{person}{Peilin Zhao}, \bibinfo{person}{Steven~CH Hoi}, \bibinfo{person}{Jialei Wang}, {and} \bibinfo{person}{Bin Li}.} \bibinfo{year}{2014}\natexlab{}.
\newblock \showarticletitle{Online transfer learning}.
\newblock \bibinfo{journal}{\emph{Artificial intelligence}}  \bibinfo{volume}{216} (\bibinfo{year}{2014}), \bibinfo{pages}{76--102}.
\newblock


\bibitem[Zhuang et~al\mbox{.}(2020)]%
        {zhuang2020comprehensive}
\bibfield{author}{\bibinfo{person}{Fuzhen Zhuang}, \bibinfo{person}{Zhiyuan Qi}, \bibinfo{person}{Keyu Duan}, \bibinfo{person}{Dongbo Xi}, \bibinfo{person}{Yongchun Zhu}, \bibinfo{person}{Hengshu Zhu}, \bibinfo{person}{Hui Xiong}, {and} \bibinfo{person}{Qing He}.} \bibinfo{year}{2020}\natexlab{}.
\newblock \showarticletitle{A comprehensive survey on transfer learning}.
\newblock \bibinfo{journal}{\emph{Proc. IEEE}} \bibinfo{volume}{109}, \bibinfo{number}{1} (\bibinfo{year}{2020}), \bibinfo{pages}{43--76}.
\newblock


\end{thebibliography}
